\documentclass[letterpaper]{article} 
\usepackage{aaai2026}  
\usepackage{times}  
\usepackage{helvet}  
\usepackage{courier}  
\usepackage[hyphens]{url}  
\usepackage{graphicx} 
\usepackage{natbib}  
\usepackage{caption} 
\usepackage{algorithm}
\usepackage{algorithmic}
\usepackage{multirow}
\usepackage{subcaption}
\usepackage{makecell}
\usepackage{amsmath}
\usepackage{amssymb}
\usepackage{amsthm}
\usepackage{booktabs}
\usepackage{multirow}
\usepackage{colortbl}
\usepackage[table]{xcolor}
\usepackage{caption}
\usepackage{subcaption}
\usepackage{bm}
\usepackage{siunitx}
\usepackage{tcolorbox}
\usepackage{booktabs}

\newtheorem{theorem}{Theorem}
\newtheorem{corollary}{Corollary}
\newtheorem{lemma}{Lemma}

\usepackage{times}
\usepackage{helvet}
\usepackage{courier}
\usepackage{xcolor}

\usepackage{newfloat}
\usepackage{listings}
\DeclareCaptionStyle{ruled}{labelfont=normalfont,labelsep=colon,strut=off} 
\floatstyle{ruled}
\newfloat{listing}{tb}{lst}{}
\floatname{listing}{Listing}
\title{Catastrophic Forgetting in Kolmogorov-Arnold Networks}
\author {
    Mohammad Marufur Rahman\textsuperscript{\rm 1},
    Guanchu Wang\textsuperscript{\rm 2},
    Kaixiong Zhou\textsuperscript{\rm 3},
    Minghan Chen\textsuperscript{\rm 1},
    Fan Yang\textsuperscript{\rm 1}
}
\affiliations {
    \textsuperscript{\rm 1}Department of Computer Science, Wake Forest University\\
    \textsuperscript{\rm 2}Department of Computer Science, University of North Carolina at Charlotte\\
    
     \textsuperscript{\rm 3}Department of Electrical and Computer Engineering, North Carolina State University\\
    rahmm224@wfu.edu, gwang16@charlotte.edu, kzhou22@ncsu.edu, chenm@wfu.edu, yangfan@wfu.edu
}
\usepackage{bibentry}

\begin{document}

\maketitle

\begin{abstract}
Catastrophic forgetting is a longstanding challenge in continual learning, where models lose knowledge from earlier tasks when learning new ones. While various mitigation strategies have been proposed for Multi-Layer Perceptrons (MLPs), recent architectural advances like Kolmogorov-Arnold Networks (KANs) have been suggested to offer intrinsic resistance to forgetting by leveraging localized spline-based activations. However, the practical behavior of KANs under continual learning remains unclear, and their limitations are not well understood. To address this, we present a comprehensive study of catastrophic forgetting in KANs and develop a theoretical framework that links forgetting to \textit{activation support overlap} and \textit{intrinsic data dimension}. We validate these analyses through systematic experiments on synthetic and vision tasks, measuring forgetting dynamics under varying model configurations and data complexity. Further, we introduce KAN-LoRA, a novel adapter design for parameter-efficient continual fine-tuning of language models, and evaluate its effectiveness in knowledge editing tasks. Our findings reveal that while KANs exhibit promising retention in low-dimensional algorithmic settings, they remain vulnerable to forgetting in high-dimensional domains such as image classification and language modeling. These results advance the understanding of KANs’ strengths and limitations, offering practical insights for continual learning system design.  
\end{abstract}

\begin{links}
    \link{Code}{https://github.com/marufur-cs/AAAI26}
\end{links}

\section{Introduction}

Catastrophic forgetting, also known as catastrophic interference \cite{MCCLOSKEY1989109}, a fundamental challenge in machine learning, occurs when a neural network loses previously acquired information while learning from new data. This phenomenon is central to the field of continual learning, where models are trained incrementally on non-stationary data distributions \cite{vandeven2024continuallearningcatastrophicforgetting, kemker2017measuringcatastrophicforgettingneural}. Moreover, it is prevalent in a wide range of research fields such as meta-learning \cite{spigler2020metalearntpriorsslowcatastrophic}, domain adaptation \cite{xu2020forgetnotreducingcatastrophic}, foundation models \cite{luo2025empiricalstudycatastrophicforgetting}, and reinforcement learning \cite{9760159}, where the retention of prior knowledge is critical for generalization and stability. 

Multi-Layer Perceptrons (MLPs) are inherently prone to catastrophic forgetting \cite{PMID:31909397}. Several techniques have been proposed to overcome catastrophic forgetting in MLPs \cite{10752992, 9349197}. Regularization-based techniques \cite{Kirkpatrick_2017, 10190202} impose restrictions on the network's weight adjustments, hence reducing the likelihood of interference with previously acquired knowledge. Architecture-based methods \cite{yoon2018lifelonglearningdynamicallyexpandable, mirzadeh2022architecturematterscontinuallearning} mitigate forgetting by modifying the network's architecture to accommodate new information. Rehearsal-based methods \cite{buzzega2020darkexperiencegeneralcontinual, riemer2019learninglearnforgettingmaximizing} aim to preserve prior information by including data samples from earlier learning sessions during the current session. Although catastrophic forgetting has been extensively studied in MLPs, it remains relatively underexplored in emerging fundamental neural architectures such as Kolmogorov-Arnold Networks (KANs) \cite{liu2025kankolmogorovarnoldnetworks}. 

KANs, inspired by the  Kolmogorov-Arnold representation theorem~\cite{kolmogorov1961representation}, have emerged as a promising alternative neural network architecture to traditional MLPs. KANs were introduced to address several fundamental limitations of MLPs. Unlike MLPs, which rely on fixed activation functions, KANs utilize learnable one-dimensional activation functions (spline) along the edges of the network. Splines can be easily adjusted locally and are accurate for low-dimensional functions, giving KANs the potential to avoid forgetting. As spline bases are local, a data sample affects only a few related spline coefficients, leaving other coefficients unaltered. This unique architecture enables KANs to learn non-linear relations more effectively and to be more robust against catastrophic forgetting in continual learning scenarios \cite{Lee_Gomes_Zhang_Kleijn_2025}. KANs have been successfully applied in various domains \cite{yang2024kolmogorovarnoldtransformer,10681070}, yet studies around their effectiveness in mitigating catastrophic forgetting in continual learning are still quite limited. 

Only a few pioneer works have studied the catastrophic forgetting phenomenon in KANs under the continual learning settings. \citeauthor{Lee_Gomes_Zhang_Kleijn_2025} recently proposed a simple and heuristic strategy, WiseKAN, which allocates distinct parameter subspaces to different tasks to mitigate catastrophic forgetting in KANs. \citeauthor{liu2025kankolmogorovarnoldnetworks} demonstrated robustness of KANs against catastrophic forgetting using synthetic data on regression tasks. Furthermore, some studies proposed modified KANs to achieve robust retention in specific domains, such as classification \cite{hu2025kackolmogorovarnoldclassifiercontinual} and face forgery detection \cite{zhang2025unifyinglocalitykansfeature} tasks. Despite these initial efforts, a comprehensive understanding of forgetting in KANs remains elusive, particularly in terms of theoretical characterization and empirical evaluation on practical real-world tasks.

To bridge the gap, we first develop a theoretical framework for understanding catastrophic forgetting in KANs by formulating several key factors such as \textit{activation support overlap} and \textit{intrinsic data dimension}. Our analysis reveals that forgetting in KANs scales linearly with activation support overlap and grows exponentially with the intrinsic dimensionality of the task manifold, offering a principled explanation for KANs’ robustness in simple tasks and vulnerability in complex domains. Building on these insights, we then conduct extensive empirical experiments comparing KANs with MLPs across a spectrum of tasks, including the low-dimensional synthetic addition and the high-dimensional image classification. Furthermore, we design a novel LoRA~\cite{hu2022lora} adapter based on KAN, termed KAN-LoRA, to enable continual fine-tuning of language models (LMs) for sequential knowledge editing. Across all experimental settings, our results consistently corroborate the theoretical analysis, illustrating that while KANs achieve strong retention in structured and low-dimensional tasks, they remain susceptible to forgetting in high-dimensional domains, thereby highlighting both the strengths and limitations of KANs in practical continual learning scenarios. Our main contributions are summarized as below: 
 
\begin{itemize}
\item We develop a theoretical framework for catastrophic forgetting in KANs, deriving formal retention bounds based on activation support overlap and intrinsic data dimension, and characterizing how forgetting evolves; 

\item We validate the theoretical analysis through empirical experiments on synthetic and image data, demonstrating strong alignment between the support overlap, task complexity, and the observed forgetting behavior; 

\item We introduce KAN-LoRA, a novel KAN-based adapter for continual fine-tuning of LMs, and evaluate its performance in sequential knowledge editing, highlighting both the strength and limitations of KANs in practice. 
\end{itemize}

\section{Preliminary}

\subsection{Catastrophic Interference}
Neural networks learn the non-linear mapping between input and output spaces by finding a region in the parameter space where the network achieves expected behavior~\cite{bishop1994neural}. When the neural network is trained on new data, the network's parameter space shifts accordingly to capture the mapping between new input and output space. As a result, performance degrades on prior data. This phenomenon was termed as \textit{catastrophic interference} by~\citeauthor{MCCLOSKEY1989109}. It was observed in many machine learning models such as support vector machine \cite{cfinsvm}, but is particularly pronounced in connectionist models (e.g., MLPs) due to their dense and globally updated parameterizations~\cite{FRENCH1999128}. Standard neural training algorithms typically lack the capacity to progressively learn new tasks without overwriting previous knowledge~\cite{aleixo2023catastrophicforgettingdeeplearning}, making them especially vulnerable to catastrophic interference. Such limitation has motivated continual learning studies~\cite{9349197} to develop algorithms and architectures that enable models to acquire new knowledge incrementally while preserving performance on learned tasks.

\subsection{Kolmogorov-Arnold Networks}
KANs are inspired by the Kolmogorov-Arnold representation theorem, which states that a finite sum of continuous univariate functions and the binary addition operation can represent any multivariate continuous function $f(\mathbf{x})$ in a specified bounded domain~\cite{kolmogorov1961representation}. Based on the theorem, function $f(\mathbf{x})$ can be represented as 
\begin{equation}
    f(\mathbf{x}) = f(x_1, x_2,..., x_n) = \sum^{2n+1}_{q=1} \Psi_q(\sum^n_{p=1}\psi_{p,q}(x_p)), 
    \label{eq:kan1}
    \nonumber
\end{equation}
where $n$ is the number of input variables, $\psi_{p,q} : \left[0,1\right] \xrightarrow{} \mathbb{R}$, and $\Psi_q:\mathbb{R}\xrightarrow{} \mathbb{R}$. 
This equation indicates that a 2-layer network with $n$ inputs and $(2n+1)$ outputs is sufficient to represent $f(\mathbf{x})$ by the sums of univariate functions. However, 1-D function $\psi$ can be fractal and non-smooth, making it unlearnable \cite{kanfails} in practice. KANs solve this issue by generalizing the theorem to multiple layers with arbitrary width. Formally, KANs consisting $L$ layers can be indicated by
\begin{equation}
    f(x) = (\Phi_{L-1} \circ \Phi_{L-2} \circ \cdots \circ \Phi_1 \circ \Phi_0)(x),
    \nonumber
\end{equation}

\begin{equation}
\Phi_\ell =
\begin{bmatrix}
\phi_{\ell,1,1} & \phi_{\ell,2,1} & \cdots & \phi_{\ell,d_{\ell},1} \\
\phi_{\ell,1,2} & \phi_{\ell,2,2} & \cdots & \phi_{\ell,d_{\ell},2} \\
\vdots     & \vdots     & \ddots & \vdots \\
\phi_{\ell,1,N_{\ell}} & \phi_{\ell,2,N_{\ell}} & \cdots & \phi_{\ell,d_{\ell},N_{\ell}}
\end{bmatrix},
\nonumber
\end{equation}
where $\circ$ indicates matrix multiplication, $\Phi_\ell$ is the function matrix that corresponds to the $\ell$-th layer, $d_{\ell}$ and $N_{\ell}$ are the number of input coordinates and univariate branches respectively. The univariate function $\phi$ is defined as the weighted sum of a base and a spline function~\cite{liu2025kankolmogorovarnoldnetworks}.

\section{Forgetting Analysis} 
\label{sec:theory}

We first introduce a formal measure of \textit{forgetting} and the notation needed to analyze how KAN’s local activations give rise to both perfect retention and task interference. 
Let $f^{(t)}$ denote the KAN obtained after sequentially training on tasks $1,2,\dots,t$, and define
\[
F_i = L\bigl(f^{(T)},\mathcal D_i\bigr) - L\bigl(f^{(i)},\mathcal D_i\bigr)
\]
as the \emph{forgetting} on task $i$, where $L(f,\mathcal D)$ is the expected loss under data distribution $\mathcal D$. We index layers by $\ell\in\{1,\dots,L\}$, and within each layer we number the input coordinates (pre-activations) by $p\in\{1,\dots,d_\ell\}$ and the individual univariate branches by $q\in\{1,\dots,N_\ell\}$.

To capture where each unit actually “turns on”, we define the \emph{activation support} of branch $\phi_{\ell,p,q}$ for task $i$ as
\[
S^{(i)}_{\ell,p,q} = \{\,z\in\mathbb R : \phi_{\ell,p,q}(z)\neq0\},
\]
the subset of real inputs on which that branch contributes non-zero output. We measure the size of these one-dimensional sets by the Lebesgue measure $\mu(\cdot)$\footnote{Lebesgue measure generalizes the length to a broader class of sets. Here, it corresponds to the total length of the activation region.}. With these setups, we can represent the maximum one-dimensional overlap between any single activation for tasks $i$ and $j$ as
\[
\Delta_{i,j} = \max_{\ell,p,q}\;\mu\bigl(S^{(i)}_{\ell,p,q}\cap S^{(j)}_{\ell,p,q}\bigr),
\]
which will serve as the key link between KAN’s architectural locality and the bounds on catastrophic forgetting. \footnote{Detailed derivations for all theorems are in the full version.}

\subsection{Bounded Retention}

We now precisely characterize when KAN achieves perfect retention and how any residual overlap translates into bounded forgetting. Overall, we demonstrate that KAN’s local‐support activations act as task‐specific feature detectors: if their “on” regions never coincide across tasks, earlier knowledge remains untouched, and when they do overlap, forgetting grows in direct proportion to that overlap.
\begin{lemma}[Zero‐Overlap Retention]\label{lem:zero_overlap}
Suppose for an earlier task $i$ and every later task $j>i$ the maximal support‐overlap satisfies $\Delta_{i,j}=0$. Then
\[
F_i
\;=\;
L\bigl(f^{(T)},\mathcal D_i\bigr)
\;-\;
L\bigl(f^{(i)},\mathcal D_i\bigr)
\;=\;
0.
\]
\end{lemma}
\begin{tcolorbox}[colback=gray!10!white,colframe=gray!75!black,boxrule=1pt,title=Remark on Lemma~\ref{lem:zero_overlap}]
When no branch ever activates on both tasks, gradient updates for new tasks cannot affect the parameters responsible for task $i$, guaranteeing \emph{exact} retention. This lemma formalizes the intuition that \textbf{truly disjoint representations cannot interfere}.
\end{tcolorbox}

\begin{theorem}[Retention Bound via Overlap]\label{thm:retention_bound}
Under the additional assumptions that each branch $\phi_{\ell,p,q}$ is $L_\ell$-Lipschitz\footnote{$\phi$ is $L$-Lipschitz if $|\phi(z_1) - \phi(z_2)| \le L|z_1 - z_2|$ for all $z_1, z_2 \in \mathbb{R}$. Here, $L_\ell$ quantifies the spline smoothness in layer $\ell$.} and the loss is bounded by $C$, for any $j>i$ we have
\[
F_i
\;\le\;
C
\sum_{\ell=1}^L
N_\ell\,L_\ell\;\Delta_{i,j}.
\]
\end{theorem}
\begin{tcolorbox}[colback=gray!10!white,colframe=gray!75!black,boxrule=1pt,title=Remark on Theorem~\ref{thm:retention_bound}]
This bound reveals that any forgetting in KANs \textbf{scales linearly with the one‐dimensional overlap} $\Delta_{i,j}$ and the network’s size parameters. Importantly, when $\Delta_{i,j}=0$, it collapses to $F_i\le0$, recovering exact retention as a special case and showing that small overlaps incur proportionally small forgetting.
\end{tcolorbox}

\subsection{Cumulative Forgetting}

While Theorem~\ref{thm:retention_bound} guarantee zero or bounded forgetting on a per‐task basis, real continual learning involves sequences of overlapping tasks whose supports may intersect in complex ways. To capture the deeper dynamics of forgetting in KANs, we further analyze at the branch level and consider cumulative contributions and effects.   

\begin{theorem}[Branch‐wise Cumulative Forgetting]\label{thm:branchwise_forgetting}
Under the Lipschitz and bounded‐loss assumptions of Theorem~\ref{thm:retention_bound}, the forgetting on task \(i\) after training on all subsequent tasks \(i+1,\dots,T\) can be decomposed as
\[
F_i \;\le\; C \sum_{\ell=1}^L \sum_{p=1}^{d_\ell} \sum_{q=1}^{N_\ell}
L_\ell \;\Bigl[\sum_{j=i+1}^T \mu\bigl(S^{(i)}_{\ell,p,q}\cap S^{(j)}_{\ell,p,q}\bigr)\Bigr].
\]
\end{theorem}
\begin{tcolorbox}[colback=gray!10!white,colframe=gray!75!black,boxrule=1pt,title=Remark on Theorem~\ref{thm:branchwise_forgetting}]
Forgetting in KANs is driven not only by the largest single overlap but also by the \textbf{total overlap each branch experiences across tasks}. Branches with frequent cross‐task activation contribute disproportionately to forgetting, suggesting that sparsifying or diversifying supports could mitigate interference.
\end{tcolorbox}

\begin{corollary}[Expected Forgetting under Random Supports]\label{coro:expected_forgetting}
If each branch’s supports for task \(j\) are independently drawn as length-\(s_j\) intervals in \([0,1]\), then in expectation  
\[
\mathbb E[F_i]\;\le\;C\sum\nolimits_{\ell=1}^L N_\ell L_\ell \ \sum\nolimits_{j=i+1}^T s_i\,s_j.
\]
\end{corollary}
\begin{tcolorbox}[colback=gray!10!white,colframe=gray!75!black,boxrule=1pt,title=Remark on Corollary~\ref{coro:expected_forgetting}]
Forgetting in KANs grows with the pairwise products of support sizes: a \textbf{difficult task} (\textit{with large \(s_j\)}) can retroactively erode performance on earlier tasks, and \textbf{longer task sequences} amplify this effect. 
\end{tcolorbox}

\begin{corollary}[Saturation via Union‐Bound]\label{coro:saturation}
Let  
\[
U^{(i)}_{\ell,p,q}
=\bigcup_{j=i+1}^T\bigl(S^{(i)}_{\ell,p,q}\cap S^{(j)}_{\ell,p,q}\bigr)
\]
be the union of all overlaps for branch \((\ell,p,q)\). Then  
\[
F_i
\;\le\;
C \sum_{\ell=1}^L \sum_{p=1}^{d_\ell} \sum_{q=1}^{N_\ell}
L_\ell\;\mu\bigl(U^{(i)}_{\ell,p,q}\bigr),
\]
with \(\mu(U^{(i)}_{\ell,p,q})\le\min\bigl(\sum_{j=i+1}^T\Delta_{i,j},\,\mu(S^{(i)}_{\ell,p,q})\bigr)\).
\end{corollary}
\begin{tcolorbox}[colback=gray!10!white,colframe=gray!75!black,boxrule=1pt,title=Remark on Corollary~\ref{coro:saturation}]
Forgetting in KANs will \textbf{saturate}, once a branch’s \textbf{full activation support is covered by overlaps}. After enough highly overlapping tasks, further tasks cannot increase forgetting beyond that support size. 
\end{tcolorbox}

\subsection{Complexity-Induced Forgetting}

Beyond mere pairwise overlap, we further conduct theoretical analysis by examining how intrinsic task complexity drives forgetting in KANs. In particular, we show that when tasks live on data manifolds of differing intrinsic dimensions, the degree of forgetting can change dramatically. This complements our earlier results by linking forgetting directly to geometric measures of task difficulty.

\begin{theorem}[Intrinsic‐Dimension Forgetting Rate]\label{thm:intrinsic_dimension}
Suppose task \(t\) generates data concentrated on a compact submanifold \(\mathcal M_t\subset[0,1]^n\) of intrinsic dimension \(d_t\), and each univariate branch’s activation support can be enclosed within an \(r\)-ball in the pre-activation domain. Then, for any earlier task \(i\) and later task \(j\), the expected support overlap satisfies
\[
\mathbb E\bigl[\mu(S^{(i)}_{\ell,p,q}\!\cap\!S^{(j)}_{\ell,p,q})\bigr]
= O\bigl(r^{\,d_i + d_j}\bigr), 
\]
and hence the forgetting on task \(i\) obeys
\[
F_i \;=\; O\!\Bigl(\sum_{j=i+1}^T N_{\mathrm{tot}}\,\bar L\;r^{\,d_i + d_j}\Bigr),
\]
where \(N_{\mathrm{tot}}=\sum_\ell N_\ell\) counts the total number of univariate branches and \(\bar L\) is an average Lipschitz constant.
\end{theorem}
\begin{tcolorbox}[colback=gray!10!white,colframe=gray!75!black,boxrule=1pt,title=Remark on Theorem~\ref{thm:intrinsic_dimension}]
Tasks with \textbf{higher intrinsic dimension produce exponentially smaller “gaps”} in their activation partitions, so even modest support radius \(r\) could incur large overlaps and thus substantial forgetting. Conversely, low-dimensional tasks enjoy near-zero overlap and stable retention in continuous learning. 
\end{tcolorbox}

\begin{corollary}[Retention for Low‐Dimensional Tasks]\label{coro:robust_retention}
If every subsequent task \(j\) has intrinsic dimension \(d_j\le D\), then
\[
F_i = O\bigl(T\,N_{\mathrm{tot}}\,\bar L\;r^{\,d_i + D}\bigr),
\]
which becomes negligible when \(d_i + D\) is sufficiently small.
\end{corollary}
\begin{tcolorbox}[colback=gray!10!white,colframe=gray!75!black,boxrule=1pt,title=Remark on Corollary~\ref{coro:robust_retention}]
When both the original task and all new tasks inhabit \textbf{low-dimensional manifolds}, their \textbf{activation overlaps shrink exponentially} in dimension, protecting against forgetting even over long task sequences. 
\end{tcolorbox}

\begin{corollary}[Fragmentation Mitigates Complexity]\label{coro:fragmentation}
If each branch’s support for task \(t\) is split into \(k_t\) disjoint intervals (effective radius \(r/k_t\)), then Theorem~\ref{thm:intrinsic_dimension}’s rate improves to
\[
F_i = O\!\Bigl(\sum_{j=i+1}^T N_{\mathrm{tot}}\,\bar L\;\bigl(r/k_i\bigr)^{d_i}\bigl(r/k_j\bigr)^{d_j}\Bigr).
\]
\end{corollary}
\begin{tcolorbox}[colback=gray!10!white,colframe=gray!75!black,boxrule=1pt,title=Remark on Corollary~\ref{coro:fragmentation}]
KANs can \textbf{sharply reduce forgetting} on high-dimensional tasks \textbf{by increasing support fragmentation}, which effectively refines each branch’s receptive field and trades off coarser representation granularity for higher retention fidelity. 
\end{tcolorbox}

Overall, Theorem~\ref{thm:intrinsic_dimension} and its corollaries illuminate how KAN’s forgetting depends on the deeper geometric complexity of task data and the combinatorial structure of activation supports. This perspective provides actionable guidance for designing KAN architectures and pruning strategies.

\section{Experiments}

We conduct a series of experiments to empirically validate our theoretical findings and assess KANs’ forgetting behavior across diverse settings. Starting with low-dimensional synthetic tasks, we analyze retention under binary and decimal addition. We then evaluate KANs on high-dimensional image classification benchmarks and finally test KAN-LoRA for continual knowledge editing in LMs. These experiments effectively illustrate how model architecture and task complexity shape the forgetting in KANs. \footnote{More details on the experiments are in the full version.}

\subsection{Binary and Decimal Addition}
\subsubsection{Experimental Setup}
We construct five synthetic tasks under a continual setting. Each task is defined by fixing one of the operands in a two-digit addition problem. Specifically, Task 1 involves \textit{one's addition}, where the digit $1$ is added to every digit from $1$ to $9$. Task 2 is \textit{two's addition}, and so forth up to \textit{five's addition} in Task 5. We apply this construction for both binary and decimal representations of digits. This setup enables us to systematically evaluate forgetting across increasingly overlapping arithmetic patterns. 
\begin{figure}[t]
    \centering
        \includegraphics[width=0.75\linewidth]{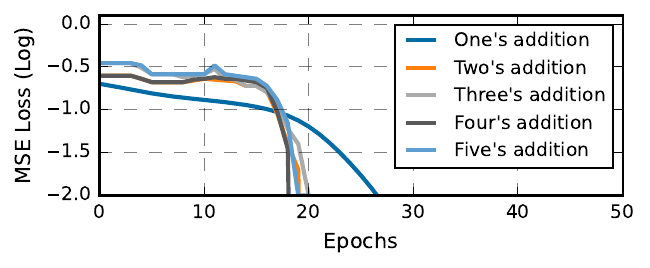}
        \caption{MSE loss in logarithmic scale for five different binary addition tasks during training on \textit{one's addition} task.}
        \label{fig:binary}
\end{figure}
\begin{figure*}[t]
    \centering
        \includegraphics[width=0.90\linewidth]{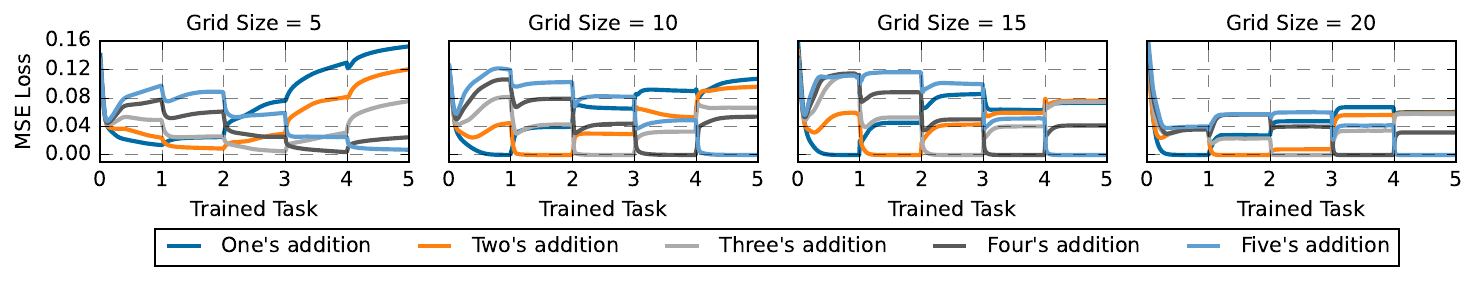}
    \caption{MSE loss during sequential training on five different decimal addition tasks, from \textit{one's} to \textit{five's addition} facts.}
    \label{fig:decimal}
\end{figure*}

\subsubsection{Model Configuration}
Our KAN model is configured with three input nodes, two hidden neurons, and two output neurons to perform addition of two 4-bit binary numbers. At each step, the model receives two input bits (one from each number) along with a carry bit, and outputs the corresponding sum bit and the carry bit for the next step. The univariate functions in KANs are modeled using B-splines~\cite{prautzsch2002bezier}, where the grid size determines the number of intervals in the spline. A larger grid size provides greater flexibility, allowing the splines to capture more complex functions \cite{liu2025kankolmogorovarnoldnetworks}. For binary addition tasks, we use a grid size of 5. For decimal addition tasks, we evaluate KANs with grid sizes of 5, 10, 15, and 20. As a baseline, we compare the KAN’s performance on binary addition to a specialized MLP architecture \cite{binaryaddition} designed to learn binary addition rules in a continual learning setting without catastrophic forgetting. 

\subsubsection{Evaluation Results.} The KAN model is sequentially trained on five binary addition tasks. Figure~\ref{fig:binary} shows the Mean Squared Error (MSE) loss (log scale) for all five tasks during training on the \textit{one’s addition} task. Notably, even during training on the first task, the losses for subsequent tasks also decrease significantly, indicating strong positive correlation. As training progresses, the model maintains stable performance on earlier tasks, with overall forgetting remaining below \SI{1e-6} after all five sessions, showing a strong resilience to catastrophic forgetting. The KAN outperforms the specialized MLP model designed for binary addition. While the MLP requires sequential training on both \textit{one’s} and \textit{two’s addition} tasks to succeed, the KAN model generalizes effectively after learning just the \textit{one’s addition} task.

Similarly, for the decimal addition, the KAN is trained sequentially over five tasks. Unlike the binary setting, the model does not fully learn subsequent tasks during training on the first one. As training progresses, clear signs of catastrophic forgetting emerge. Figure~\ref{fig:decimal} shows that learning a new task leads to a noticeable decline in performance on previous tasks. However, the severity of forgetting decreases as the grid size increases, suggesting that finer spline resolution improves retention. After completing all five tasks, a clear forgetting pattern appears: performance deteriorates more significantly for tasks that are farther in time from the most recent training, indicating that earlier tasks suffer more from forgetting. These observations empirically support our analysis in Corollary~\ref{coro:expected_forgetting}. On one hand, increasing the grid size reduces each spline's support length, thereby decreasing pairwise overlaps and mitigating the forgetting. On the other hand, later tasks, which have larger effective support sizes due to increased digit variability, lead to greater cumulative interference, consistent with the $s_i s_j$ dependence. 

Tables~\ref{tab:theo1} and~\ref{tab:theo2} further present empirical evidences supporting Theorems~\ref{thm:retention_bound} and~\ref{thm:branchwise_forgetting}. In Table~\ref{tab:theo1}, for each pair of tasks selected from the five decimal addition tasks, the ratio $F_i/\Delta_{i,j}$ remains approximately \textit{constant}, suggesting that the forgetting $F_i$ scales linearly with the support overlap $\Delta_{i,j}$ between tasks $i$ and $j$. Similarly, Table~\ref{tab:theo2} shows that the ratio between the observed forgetting and the cumulative support overlap\footnote{Simplify to $\sum\mu^{ij}$ in Table~\ref{tab:theo2} notations.} $\sum^T_{i+1}\mu(S^{(i)}\cap S^{(j)})$ is also nearly \textit{constant}, indicating a linear dependence. Additionally, this ratio becomes more stable (i.e., exhibits lower variance) as the grid size of KANs increases, revealing that finer-grained spline meshes promote more consistent forgetting behavior.

\begin{table}[ht]
    \centering
        \resizebox{0.99\columnwidth}{!}{%
        \begin{tabular}{c|c|cc|cc|cc}
            \toprule
            \multirow{2}{*}{\makecell{\textbf{Task ($i$)}}} & 
            \multirow{2}{*}{\makecell{\textbf{Task ($j$)}}} & 
            \multicolumn{2}{c|}{\textbf{Grid 10}} & 
            \multicolumn{2}{c|}{\textbf{Grid 15}} & 
            \multicolumn{2}{c}{\textbf{Grid 20}} \\
            \cmidrule{3-8}
            & & \bm{$F_i$} & \bm{$\frac{F_i}{\Delta_{i,j}}$} & \bm{$F_i$} & \bm{$\frac{F_i}{\Delta_{i,j}}$} & \bm{$F_i$} & \bm{$\frac{F_i}{\Delta_{i,j}}$} \\
            \midrule
            1 & 2 & 0.46 & \textbf{0.74} & 0.45 & \textbf{0.74} & 0.32 & \textbf{0.61} \\
            2 & 3 & 0.45 & \textbf{0.73} & 0.40 & \textbf{0.67} & 0.34 & \textbf{0.64} \\
            3 & 4 & 0.52 & \textbf{0.77} & 0.46 & \textbf{0.74} & 0.32 & \textbf{0.63} \\
            4 & 5 & 0.44 & \textbf{0.72} & 0.42 & \textbf{0.68} & 0.32 & \textbf{0.64} \\
            \bottomrule
        \end{tabular}
        }

    \caption{Retention bounds across KANs and tasks.}
    \label{tab:theo1}
\end{table}

{\small
\begin{table}[ht]
    \centering
        \resizebox{0.99\columnwidth}{!}{%
        \begin{tabular}{c|c|cc|cc|cc}
            \toprule
            \multirow{2}{*}{\makecell{\textbf{Task ($i$)}}} & 
            \multirow{2}{*}{\makecell{\textbf{Task ($j$)}}} & 
            \multicolumn{2}{c|}{\textbf{Grid 10}} & 
            \multicolumn{2}{c|}{\textbf{Grid 15}} & 
            \multicolumn{2}{c}{\textbf{Grid 20}} \\
            \cmidrule{3-8}
            & & \bm{$F_i$} & \bm{$\frac{F_i}{\sum\mu^{ij}}$} & \bm{$F_i$} & \bm{$\frac{F_i}{\sum\mu^{ij}}$} & \bm{$F_i$} & \bm{$\frac{F_i}{\sum\mu^{ij}}$} \\
            \midrule
            1 & 2, 3, 4, 5 & 0.68 & \textbf{0.15} & 0.62 & \textbf{0.15} & 0.57 & \textbf{0.16} \\
            2 & 3, 4, 5 & 0.67 &\textbf{ 0.16} & 0.51 & \textbf{0.15} & 0.44 & \textbf{0.16} \\
            3 & 4, 5 & 0.39 & \textbf{0.16} & 0.39 & \textbf{0.16} & 0.29 & \textbf{0.16} \\
            4 & 5 & 0.25 & \textbf{0.18} & 0.19 & \textbf{0.17} & 0.16 & \textbf{0.17} \\
            \bottomrule
        \end{tabular}
        }

    \caption{Cumulative bounds across KANs and tasks.} 
    \label{tab:theo2}
\end{table}
}

\subsection{Image Classification}
\subsubsection{Experimental Setup}
To evaluate KANs in real-world settings, we assess their forgetting behavior with continual learning using CIFAR-10, Tiny-ImageNet, and MNIST datasets. CIFAR-10 consists $(32*32)$-pixel images of 10 evenly distributed classes. To simulate a class-incremental continual-learning scenario, the dataset is divided into five sequential tasks, each containing images from two different classes. A similar five-task setup is constructed for Tiny-ImageNet dataset, by selecting 10 classes from 200 different classes of $(64*64)$-pixel images. MNIST, $(28*28)$ pixels, is likewise divided into five sequential tasks. These three datasets vary in intrinsic dimensionality, where MNIST has the lowest while Tiny-ImageNet has the highest dimension. 

\subsubsection{Model Configuration}
We adopt a Transformer-based architecture for image classification, in which all MLP layers are replaced with KAN layers, resulting in the KAN-Transformer model~\cite{yang2024kolmogorovarnoldtransformer}. This modification intends to utilize the adaptive capacity of KANs within the Transformer framework for continual learning scenarios. To provide a fair and competitive baseline, we also design an MLP-based transformer model augmented with the EWC~\cite{Kirkpatrick_2017} regularization technique. 

\subsubsection{Evaluation Results.}
\begin{figure}[t]
    \centering
        \includegraphics[width=\linewidth]{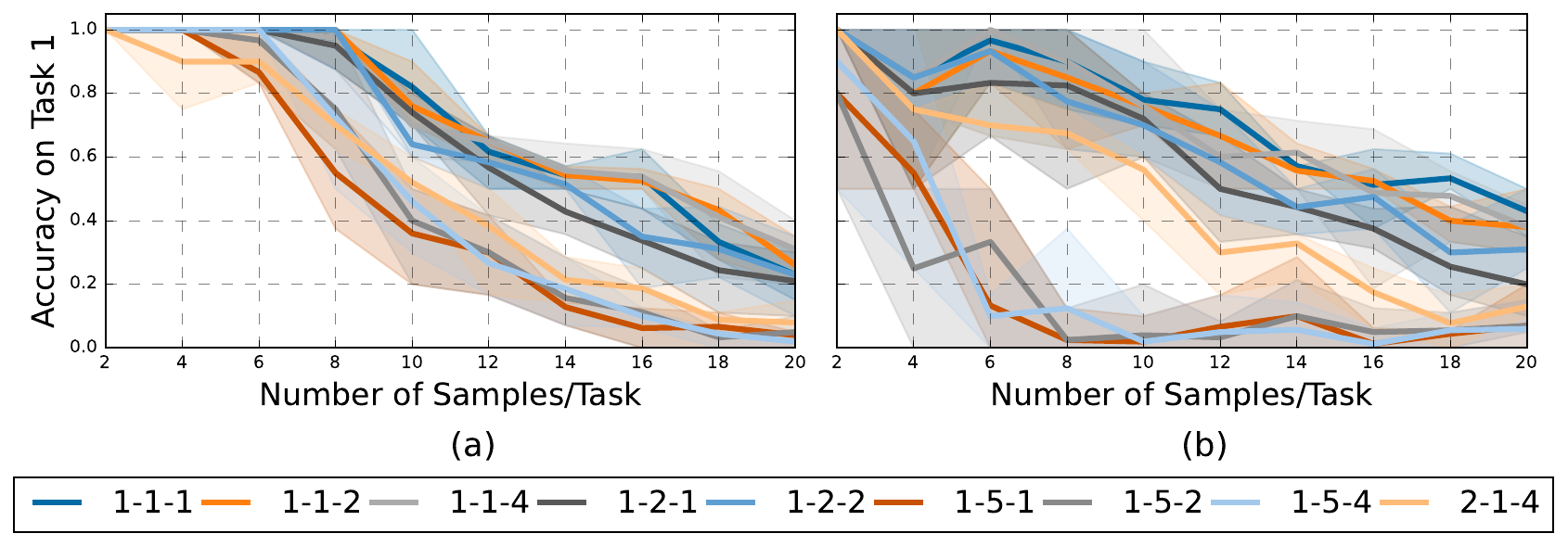}
   \caption{
Task 1 accuracy after sequential training on task 1 and 2 from CIFAR-10, comparing (a) KAN-Transformer and (b) MLP-Transformer. Model configuration is labeled as (\#classification \textit{layers} – \#encoder \textit{blocks} – \#attention \textit{heads}).
}
    \label{fig:img2tasks}
\end{figure}

Figure~\ref{fig:img2tasks} illustrates the accuracy on task 1 after sequential training of the KAN (with grid size $10$) and the MLP model on tasks 1 and 2 from CIFAR-10, evaluated across various model configurations and increasing sample sizes per task. Both architectures retain high accuracy in shallow settings with a single encoder block, attention head, and classification layer. Notably, the KAN model demonstrates superior retention, maintaining 100\% accuracy on task 1 up to 8 samples per task, whereas the MLP model drops to around 80\%. As the number of encoder blocks and classification layers increases, performance declines sharply, particularly in MLPs, which suggests deeper networks are more susceptible to catastrophic forgetting.

\begin{figure}[t]
    \centering
        \includegraphics[width=.885\linewidth]{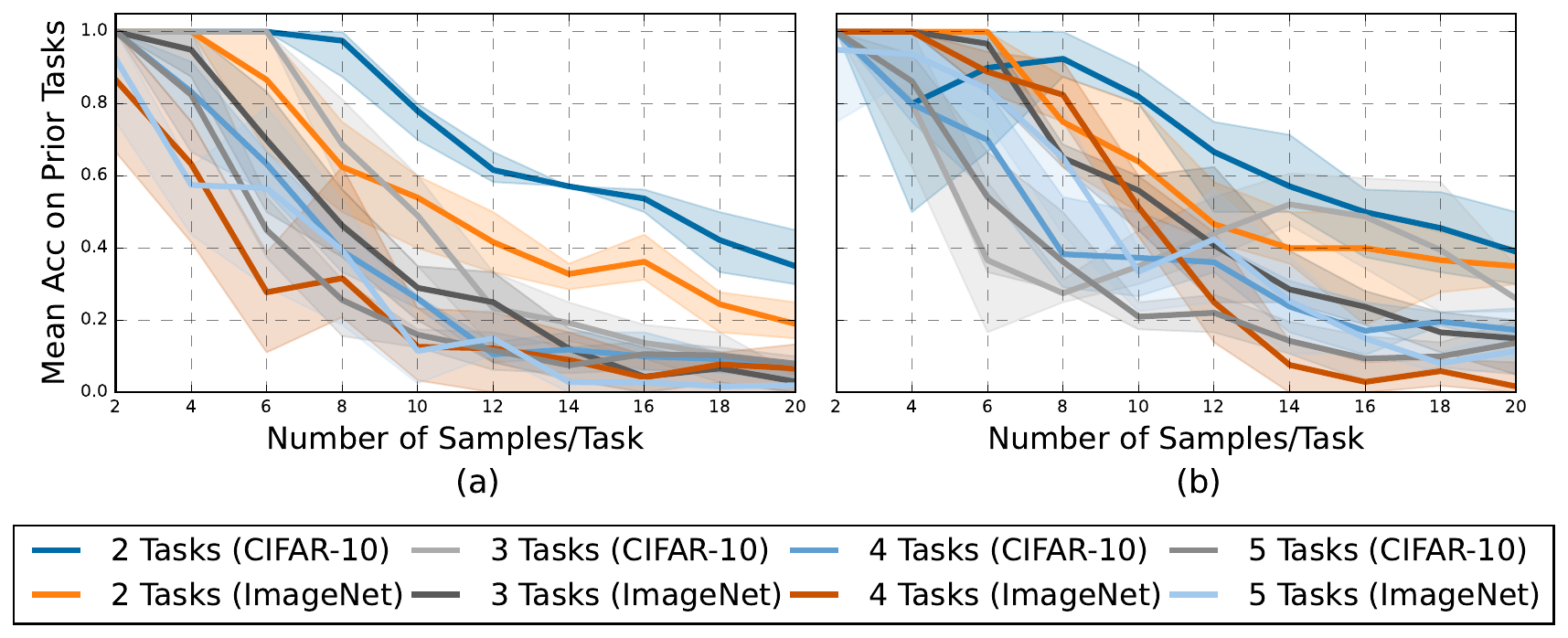}
    \caption{Average accuracy on previously learned tasks after training on 2 to 5 tasks with varying sample sizes from CIFAR-10 and Tiny-ImageNet datasets. Sub-figures show results for (a) KAN-Transformer and (b) MLP-Transformer.}
    \label{fig:img5tasks}
\end{figure}

Figure~\ref{fig:img5tasks} summarizes the impact of varying the number of samples per task during continual learning, evaluated across different task counts (ranging from 2 to 5) for both CIFAR-10 and Tiny-ImageNet. All models use one single encoder block, attention head, and classification layer. On CIFAR-10, KAN models exhibit better retention than their MLP counterparts, particularly when trained on a smaller number of tasks. In contrast, MLP models outperform KANs on the more challenging Tiny-ImageNet dataset. These results underscore the increasing difficulty of continual learning in KANs as both the number of tasks and the underlying data complexity grow. Moreover, the performance curves in Figure~\ref{fig:img5tasks}a suggest a clear saturation effect: after a certain number of highly overlapping tasks, additional training yields diminishing increases in forgetting, consistent with the bounded cumulative interference described in Corollary~\ref{coro:saturation}, where support unions eventually stabilize.

Table~\ref{tab:theo3} further presents empirical evidence supporting Theorem~\ref{thm:intrinsic_dimension}. Forgetting $F_i$ is measured on task 1 after sequential training on all five tasks from MNIST, CIFAR-10, and Tiny-ImageNet. To vary the intrinsic dimension $d_i$, the images are quantized using different label sets ($Q$) and resized to different spatial resolutions ($S$), where $d_i = \log_2(Q \times S)$. Across datasets and configurations, the ratio $\log(F_i)/d_i$ remains approximately \textit{constant}, providing strong support for the exponential relationship between forgetting and task complexity as captured by intrinsic dimension. This behavior reflects a geometric constraint where increasing intrinsic dimension entangles KANs' localized supports, highlighting the need for dimensionality-aware tuning or support fragmentation (as in Corollary~\ref{coro:fragmentation}) to sustain better retention.

\begin{table}[t]
\setlength{\tabcolsep}{2pt}  
\renewcommand{\arraystretch}{1.2}  
\huge  
    \centering
        \resizebox{0.95\columnwidth}{!}{%
        \begin{tabular}{c|c|c|c|c|c|c|c|c}
            \toprule
            \multicolumn{3}{c|}{\textbf{MNIST}} & \multicolumn{3}{c|}{\textbf{CIFAR-10}} & \multicolumn{3}{c}{\textbf{Tiny-ImageNet}}\\
            \cmidrule{1-9}
            \multirow{2}{*}{\makecell{\textbf{Quantize}\\\textbf{label ($Q$)}}} &
            \multirow{2}{*}{\textbf{Shape ($S$)}} &
            \multirow{2}{*}{\makecell{\bm{$\frac{\log(F_i)}{d_i}$}}} &
            \multirow{2}{*}{\makecell{\textbf{Quantize}\\\textbf{label ($Q$)}}} &
            \multirow{2}{*}{\textbf{Shape ($S$)}} &
            \multirow{2}{*}{\makecell{\bm{$\frac{\log(F_i)}{d_i}$}}} &
            \multirow{2}{*}{\makecell{\textbf{Quantize}\\\textbf{label ($Q$)}}} &
            \multirow{2}{*}{\textbf{Shape ($S$)}} &
            \multirow{2}{*}{\makecell{\bm{$\frac{\log(F_i)}{d_i}$}}} \\
            &&&&&&&&\\
            \midrule
            2 & 8$\times$8 & \textbf{0.074} & 8 & 8$\times$8 & \textbf{0.046} & 8 & 16$\times$16 & \textbf{0.052}\\
            2 & 16$\times$16 & \textbf{0.071} & 8 & 16$\times$16 & \textbf{0.053} & 8 & 32$\times$32 & \textbf{0.054}\\ 
            2 & 28$\times$28 & \textbf{0.074} & 8 & 32$\times$32 & \textbf{0.046} & 8 & 64$\times$64 & \textbf{0.052}\\
            4 & 28$\times$28 & \textbf{0.071} & 16 & 32$\times$32 & \textbf{0.053} & 16 & 64$\times$64 & \textbf{0.051}\\
            8 & 28$\times$28 & \textbf{0.075} & 32 & 32$\times$32 & \textbf{0.050} & 32 & 64$\times$64 & \textbf{0.049}\\
            16 & 28$\times$28 & \textbf{0.073} & 64 & 32$\times$32 & \textbf{0.048} & 64 & 64$\times$64 & \textbf{0.048}\\
            32 & 28$\times$28 & \textbf{0.075} & 128 & 32$\times$32 & \textbf{0.047} & 128 & 64$\times$64 & \textbf{0.047}\\
            \bottomrule
        \end{tabular}
        }

    \caption{Forgetting rate for varied intrinsic dimensions.}
    \label{tab:theo3}
\end{table}

\subsection{Knowledge Editing for LMs} 
\subsubsection{Experimental Setup}
LMs require continual knowledge editing to replace outdated information and integrate new facts. To evaluate the forgetting behavior of KANs and MLPs in such high-dimensional editing scenarios, five consecutive tasks are curated from the CounterFact~\cite{meng2023locatingeditingfactualassociations} and ZsRE~\cite{levy-etal-2017-zero} benchmarks. 

\begin{table*}[!t]
\centering

\begin{subtable}[t]{0.494\textwidth}
\centering
\resizebox{\textwidth}{!}{%
\begin{tabular}{|c|c|c|c|c|c|c|c|c|c|c|}
\hline
\multirow{3}{*}{\textbf{Model}} &
\multirow{3}{*}{\textbf{Dataset}} &
\multirow{3}{*}{\makecell{\textbf{Samples} \\ \textbf{per Task}}} &
\multicolumn{4}{c|}{\textbf{KAN LoRA}} &
\multicolumn{4}{c|}{\textbf{MLP LoRA}} \\
\cline{4-11}
& & &\multicolumn{4}{c|}{\textbf{\# Trained tasks}} &
\multicolumn{4}{c|}{\textbf{\# Trained tasks}}\\
\cline{4-11}
 & & & 2 & 3  & 4  & 5  & 2  & 3  & 4  & 5 \\
\hline
\multirow{8}{*}{\makecell{Llama~2-\\7B}} 
& \multirow{4}{*}{CounterFact}
& 2 &  100 & 65 & 50 & 45 & 100 & 85 & 57  & 60\\
& & 3 &  100 & 93 & 80 & 48 & 100 & 90 & 67  & 57\\
& & 4 & 100 & 88  & 78 & 53 & 100 & 90 & 65 & 42 \\
& & 5  &  100 & 88 & 80 & 57 & 100 & 98 & 77  & 66\\
\cline{2-11}
& \multirow{4}{*}{ZsRE}
& 2 & 100  & 80 & 67 & 60 & 100 & 95 &  97 & 87\\
& & 3 & 100  & 87 & 71 & 58 & 100 & 100 & 91  & 78\\
& & 4 & 100 & 85 &  70 & 55 & 100 & 82 & 63 & 46  \\
& & 5  &  100 & 76 & 64 & 60 & 100 & 86 & 73  & 57\\
\hline
\multirow{8}{*}{\makecell{Llama~2-\\13B}} 
& \multirow{4}{*}{CounterFact}
& 2 & 100  & 75 & 60 & 50 & 100 & 70 & 50 & 43\\
& & 3 & 100  & 93 & 71 & 60 & 100 & 93 & 62 & 53\\
& & 4 & 100 & 97 & 60 & 44 & 100 & 97 & 77 & 56 \\
& & 5  & 100  & 76 & 73 & 57 & 100 & 84 & 83 & 63\\
\cline{2-11}
& \multirow{4}{*}{ZsRE}
& 2 & 100  & 100 & 83 & 72 & 100 & 80 & 53  & 52\\
& & 3 &  100 & 97 & 78 & 75 & 100 & 89 & 62  & 60\\
& & 4 & 100 & 75 & 66 & 58 & 100 & 92 & 81 & 55 \\
& & 5  &  100 & 85 & 80 & 67 & 100 & 83 & 73  & 59\\
\hline
\end{tabular}
}
\caption{KAN-LoRA and MLP-LoRA adapters with rank 8.}
\label{tab:llmacc1}
\end{subtable}
\begin{subtable}[t]{0.48\textwidth}
\centering
\resizebox{\textwidth}{!}{%
\begin{tabular}{|c|c|c|c|c|c|c|c|c|c|c|}
\hline
\multirow{3}{*}{\textbf{Model}} &
\multirow{3}{*}{\textbf{Dataset}} &
\multirow{3}{*}{\makecell{\textbf{Samples} \\ \textbf{per Task}}} &
\multicolumn{4}{c|}{\textbf{KAN LoRA}} &
\multicolumn{4}{c|}{\textbf{MLP LoRA}}\\
\cline{4-11}
& & &\multicolumn{4}{c|}{\textbf{\# Trained tasks}} &
\multicolumn{4}{c|}{\textbf{\# Trained tasks}}\\
\cline{4-11}
 & & & 2 & 3  & 4  & 5  & 2  & 3  & 4  & 5 \\
\hline

\multirow{8}{*}{\makecell{Llama~2-\\7B}} 
& \multirow{4}{*}{CounterFact}

& 2 & 100  & 75 & 47 & 45 & 100 & 55 & 30  & 43\\
& & 3 & 100  & 77 & 62 & 55 & 100 & 80 & 53  & 45\\
& & 4 & 100 & 85  & 57 & 41 & 100 & 75 & 60 & 43 \\
& & 5  & 100 & 70 & 60 & 49 & 100 & 76 & 59  & 51\\
\cline{2-11}
& \multirow{4}{*}{ZsRE}
& 2 &  100 & 90 & 50 & 45 & 100 & 55 &  43 & 37\\
& & 3 &  100 & 77 & 58 & 57 & 100 & 70 & 49  & 27\\
& & 4 & 100 & 70 & 65  & 48 & 100 & 70 & 58 & 39  \\
& & 5  &  100 & 74 & 63 & 49 & 100 & 86 & 59  & 39\\
\hline

\multirow{8}{*}{\makecell{Llama~2-\\13B}} 
& \multirow{4}{*}{CounterFact}
& 2 & 100  & 65 & 50 & 60 & 100 & 55 & 43 & 40\\
& & 3 & 100  & 80 & 69 & 57 & 100 & 77 & 56 & 45\\
& & 4 & 100 & 88 & 62 & 39 & 100 & 85 & 50 & 34 \\
& & 5  & 100 & 64 & 61 & 48 & 100 & 84 & 65 & 45\\
\cline{2-11}
& \multirow{4}{*}{ZsRE}
& 2 & 100 & 70 & 53 & 45 & 100 & 55 & 47  & 40\\
& & 3 & 100  & 80 & 60 & 57 & 100 & 73 & 47  & 40\\
& & 4 & 100 & 70 & 63 & 41 & 100 & 82 & 62 & 48 \\
& & 5  & 100  & 72 & 59 & 50 & 100 & 76 & 63  & 44\\
\hline

\end{tabular}
}
\caption{KAN-LoRA and MLP-LoRA adapters with rank 16.}
\label{tab:llmacc2}
\end{subtable}


\caption{Mean accuracy (\%) on previously edited tasks during continual fine-tuning of Llama~2-7B and Llama~2-13B models equipped with KAN-LoRA and MLP-LoRA adapters. Performance is reported across five consecutive tasks for each dataset.} 
\label{tab:llmacc}
\end{table*}

\subsubsection{Model Configuration}
LoRA~\cite{hu2022lora} is a parameter-efficient fine-tuning technique that adapts LMs by freezing pre-trained weights and training lightweight adapters, substantially reducing memory usage and computational cost compared to full fine-tuning~\cite{biderman2024loralearnsforgets}. To explore the use of KAN as a LoRA adapter for continual fine-tuning, we design a modified adapter architecture. In the standard LoRA setup, the frozen weight matrix $W_0 \in \mathbb{R}^{a \times b}$ is augmented by a trainable low-rank residual matrix $\Delta W \in \mathbb{R}^{a \times b}$, factorized as $\Delta W = BA$ with $B \in \mathbb{R}^{a \times c}$ and $A \in \mathbb{R}^{c \times b}$, where rank $c \ll \min(a, b)$. The adapter’s output is $h = W_0x + BAx$. In our design, both $A$ and $B$ are parameterized using KANs. This KAN-based variant, referred as KAN-LoRA, is integrated into the final two layers of Llama2-7B and Llama2-13B~\cite{touvron2023llamaopenefficientfoundation}. We apply EWC regularization during continual fine-tuning, using the preceding task as memory. For a fair comparison, we develop an MLP-LoRA adapter with identical EWC settings. For all KAN-LoRA experiments, we use a grid size of 5 to balance capacity and efficiency.

\subsubsection{Evaluation Results}  
The modified Llama models equipped with KAN-LoRA and MLP-LoRA adapters are continually fine-tuned across multiple tasks. Tables~\ref{tab:llmacc1} and \ref{tab:llmacc2} report the mean accuracy on previously edited tasks after sequential edits of varying lengths, for adapter ranks 8 and 16 respectively, highlighting the extent of forgetting during the continual fine-tuning process. Increasing the adapter rank leads to greater forgetting in both KAN and MLP variants. However, KAN adapters consistently outperform their MLP counterparts at rank 16 and in low-sample (per task) regimes. Notably, the KAN adapter shows reduced forgetting in Llama2-13B, while the MLP adapter displays the opposite trend. In small-sample settings, KAN achieves consistently higher retention in Llama2-13B compared to MLP. These results suggest that KAN adapters are more resilient to forgetting in large-scale LMs, especially at higher ranks and under limited task supervision.

\begin{table}[tbp]
\centering
\resizebox{\columnwidth}{!}{%
\begin{tabular}{c|c|c|c|c}
\toprule
\multirow{2}{*}{\textbf{Model}} &
\multirow{2}{*}{\textbf{Adapter}} &
\multirow{2}{*}{\makecell{\textbf{Trainable} \\ \textbf{parameters}}} &
\multirow{2}{*}{\makecell{\textbf{Training} \\ \textbf{time (s/epoch)}}} &
\multirow{2}{*}{\makecell{\textbf{Inference} \\ \textbf{time (s/sample)}}}\\
&&&&\\
\cmidrule{1-5}
\multirow{2}{*}{Llama~2-7B}&
KAN LoRA &
2.6M &
0.57 &
0.13 \\
\cmidrule{2-5}
& MLP LoRA&
0.28M &
0.54 &
0.12 \\
\cmidrule{1-5}
\multirow{2}{*}{Llama~2-13B}&
KAN LoRA &
3.2M &
1.05 &
0.23 \\
\cmidrule{2-5}
& MLP LoRA&
0.35M &
1.01 &
0.21 \\
\bottomrule
\end{tabular}
}
\caption{Comparison of trainable parameters, training, and inference time for KAN-LoRA and MLP-LoRA adapters.}
\label{tab:param}
\end{table}

Table~\ref{tab:param} further compares the computational and parameter efficiency of KAN-LoRA and MLP-LoRA adapters, using a grid size of 5 and an adapter rank at 8. KAN introduces significantly more trainable parameters than MLP, approximately $10\times$ more for both Llama2-7B and Llama2-13B models. Training and inference times are measured on the CounterFact dataset with 5 samples per task. While the KAN adapter incurs higher computational cost than the MLP variant, the overhead remains moderate relative to the observed gain in model capacity and retention performance.


\section{Conclusion \& Discussion} 

In this work, we present the first comprehensive study of catastrophic forgetting in KANs under continual learning settings. We develop a theoretical framework that connects forgetting dynamics to the architectural locality of spline activations and the intrinsic dimensionality of task data. Our analysis yields formal retention bounds and characterizes the cumulative and geometry-driven nature of forgetting in KANs. To validate these insights, we conduct systematic experiments across synthetic arithmetic tasks and real-world image classification benchmarks. Empirical results strongly corroborate the our analysis, revealing a clear linear relationship between forgetting and activation overlap, and an exponential increase in forgetting as task dimensionality rises. We further introduce KAN-LoRA, a novel adapter design for continual fine-tuning of LMs in model editing tasks, and demonstrate its retention superiority compared to MLP-based alternatives. Our findings establish both the strengths and limitations of KANs for continual learning.

Stepping further, we believe this work opens up several unconventional directions for advancing KANs in continual learning. First, the evolution of spline activations across tasks suggests a \emph{new dynamic view} of learning, where forgetting reflects adaptation pressures on local function regions. Designing KANs with mechanisms that support controlled specialization or even lifecycle-based pruning of splines may enhance long-term retention. Second, our analysis of support overlap points to the possibility of \emph{distributed memory encoding}. Rather than eliminating interference, future models could intentionally overlap supports to store multiple tasks in a compressed fashion that allows task-specific retrieval through decoding strategies. Third, forgetting itself may also function as a \emph{beneficial inductive bias}. Selective decay in high-dimensional regions could suppress redundant or unstable features, reduce overfitting, and improve generalization. These visions can largely reframe forgetting not as a flaw to be eliminated, but as a property to be shaped, positioning KANs as a flexible and interpretable foundation for memory-aware continual learning systems.

\section{Acknowledgements}
The authors thank the anonymous reviewers for comments.
This research was supported by the NSF IIS2451480.

\bibliography{aaai2026}

\clearpage
\appendix
\section{A.  Lemma 1 Proof}
\label{lemma1}

Fix an earlier task \(i\) and assume that for every later task \(j>i\)
\[
S^{(i)}_{\ell,p,q}\cap S^{(j)}_{\ell,p,q}=\varnothing
\quad\forall\,(\ell,p,q).
\]

\smallskip
\noindent
Let
\[
\Theta^{(i)}
  :=\bigl\{
        \theta_{\ell,p,q}:
        \mu\!\bigl(S^{(i)}_{\ell,p,q}\bigr)>0
     \bigr\}.
\]

\noindent
For \((x,y)\sim\mathcal D_j\) with \(j>i\) and any
\(\theta_{\ell,p,q}\in\Theta^{(i)}\),
the computation path cannot traverse branch \(\phi_{\ell,p,q}\) and thus 
\[
\frac{\partial}{\partial\theta_{\ell,p,q}}
      \mathcal{L}\bigl(f(x),y\bigr)=0,
\]
where \(\mathcal L\bigl(f(x),y\bigr)\) denotes the per-example loss. 

\noindent
With update rule
\(\theta^{(t+1)}=\theta^{(t)}-\eta_t g^{(t)}\) and
\(g^{(t)}=0\) on \(\Theta^{(i)}\), we have
\[
\theta^{(T)}=\theta^{(i)}
\quad
\text{for every }\theta\in\Theta^{(i)}.
\]

\noindent
For \(x\sim\mathcal D_i\), all branches in \(\Theta^{(i)}\) are frozen,
while the rest contribute the same value as before. Thus,
\(f^{(T)}(x)=f^{(i)}(x)\).
Taking expectation over \(\mathcal D_i\) yields
\[
L\bigl(f^{(T)},\mathcal D_i\bigr)=L\bigl(f^{(i)},\mathcal D_i\bigr),
\]
and therefore
\[
F_i
  =L\bigl(f^{(T)},\mathcal D_i\bigr)
  -L\bigl(f^{(i)},\mathcal D_i\bigr)
  =0. 
\]

\section{B.  Theorem 1 Proof}
\label{theo1}

Fix an earlier task $i$ and a later task $j>i$.  
During optimization on $\mathcal D_j$ the parameters of a branch
$\phi_{\ell,p,q}$ are updated only if
$S^{(i)}_{\ell,p,q}\cap S^{(j)}_{\ell,p,q}\neq\varnothing$.
Set
\[
I_{\ell,p,q}
  :=S^{(i)}_{\ell,p,q}\cap S^{(j)}_{\ell,p,q},
  \qquad
  \mu(I_{\ell,p,q})\le\Delta_{i,j}.
\]

\smallskip
\noindent\textit{Pointwise change of one branch.}
Because both pre- and post-update splines are $L_\ell$-Lipschitz,
for every $z\in I_{\ell,p,q}$ we have
\[
\bigl|\widetilde{\phi}_{\ell,p,q}(z)-\phi_{\ell,p,q}(z)\bigr|
  \le L_\ell\mu(I_{\ell,p,q})
  \le L_\ell\Delta_{i,j}.
\tag{B-1}
\]
They coincide outside $I_{\ell,p,q}$, so (B-1) bounds their difference on
all of~$\mathbb R$.

\smallskip
\noindent\textit{Layer-wise change.}
At most $N_\ell$ branches in layer~$\ell$ overlap with task~$j$, whence
for any input $x\sim\mathcal D_i$, so we further have
\[
\Bigl|
 \sum_{q=1}^{N_\ell}
   \bigl[\widetilde{\phi}_{\ell,p,q}
        -\phi_{\ell,p,q}\bigr]
        \bigl(h^{(\ell-1)}_{p}(x)\bigr)
\Bigr|
\le N_\ell L_\ell\Delta_{i,j}.
\tag{B-2}
\]

\smallskip
\noindent\textit{Network-level change.}
Subsequent layers are unchanged, so propagating (B-2) forward yields
\[
|f^{(T)}(x)-f^{(i)}(x)|
  \le \sum_{\ell=1}^L N_\ell L_\ell\Delta_{i,j}
  \quad\forall\,x\sim\mathcal D_i.
\tag{B-3}
\]

\smallskip
\noindent\textit{Effect on the expected loss.}
Because the loss is $C$-Lipschitz in its first argument,
\[
\bigl| \mathcal{L}(f^{(T)}(x),y) - \mathcal{L}(f^{(i)}(x),y)\bigr|
  \le C\,|f^{(T)}(x)-f^{(i)}(x)|.
\]
Combining with (B-3) and taking expectation over
$(x,y)\sim\mathcal D_i$ gives
\[
F_i
  =L(f^{(T)},\mathcal D_i)-L(f^{(i)},\mathcal D_i)
  \le C\sum_{\ell=1}^L N_\ell L_\ell\Delta_{i,j},
\]
establishing the claimed bound.

\section{C.  Theorem 2 Proof}
\label{theo2}

Define the incremental loss increase for each later task $t=i+1,\dots,T$ by
\[
\delta_{i,t}
  :=L\!\bigl(f^{(t)},\mathcal D_i\bigr)
    -L\!\bigl(f^{(t-1)},\mathcal D_i\bigr).
\]
Because these increments telescope, we can derive
\[
F_i
  =
  \sum_{t=i+1}^{T}\delta_{i,t}.
  \tag{C-1}
\]

\smallskip
\noindent\textit{Bounding a single increment.}
Fix $t>i$.
A branch $\phi_{\ell,p,q}$ is updated only when
$S^{(i)}_{\ell,p,q}\cap S^{(t)}_{\ell,p,q}\neq\varnothing$.
Let
\[
I_{\ell,p,q}^{(t)}
  :=S^{(i)}_{\ell,p,q}\cap S^{(t)}_{\ell,p,q},
  \quad
  \gamma_{\ell,p,q}^{(t)}
  :=\mu\!\bigl(I_{\ell,p,q}^{(t)}\bigr).
\]
From the earlier derivation in Appendix B, updating a
$L_\ell$-Lipschitz spline on an interval of length
$\gamma_{\ell,p,q}^{(t)}$ perturbs its value by at most
$L_\ell\gamma_{\ell,p,q}^{(t)}$:
\[
\bigl|\widetilde{\phi}_{\ell,p,q}(z)-\phi_{\ell,p,q}(z)\bigr|
  \le L_\ell\gamma_{\ell,p,q}^{(t)}
  \quad(\forall z\in\mathbb R).
  \tag{C-2}
\]
Summing (C-2) over the $N_\ell$ branches of layer~$\ell$ yields
\[
\bigl|h^{(\ell)}_{\text{new}}(x)-h^{(\ell)}_{\text{old}}(x)\bigr|
  \le
  L_\ell\sum_{q=1}^{N_\ell}\gamma_{\ell,p,q}^{(t)}
  \quad(x\sim\mathcal D_i).
  \tag{C-3}
\]
Propagating (C-3) through the unchanged upper layers gives
\[
|f^{(t)}(x)-f^{(t-1)}(x)|
 \le \sum_{\ell=1}^{L}
       L_\ell\sum_{q=1}^{N_\ell}\gamma_{\ell,p,q}^{(t)}
 \quad(x\sim\mathcal D_i).
\]
Because the loss is $C$-Lipschitz,
\[
\delta_{i,t}
  \le
  C\sum_{\ell=1}^{L}
      L_\ell\sum_{q=1}^{N_\ell}
            \gamma_{\ell,p,q}^{(t)}.
  \tag{C-4}
\]

\smallskip
\noindent\textit{Accumulating all later tasks.}
Insert (C-4) into the telescope (C-1) and interchange sums:
\[
\begin{aligned}
F_i
&\le
C
\sum_{t=i+1}^{T}
\sum_{\ell=1}^{L}
\sum_{p=1}^{d_\ell}
L_\ell
\sum_{q=1}^{N_\ell}
\gamma_{\ell,p,q}^{(t)}
\\[4pt]
&=
C
\sum_{\ell=1}^{L}
\sum_{p=1}^{d_\ell}
\sum_{q=1}^{N_\ell}
L_\ell
\Bigl[
  \sum_{j=i+1}^{T}
  \mu\!\bigl(
    S^{(i)}_{\ell,p,q}\cap S^{(j)}_{\ell,p,q}
  \bigr)
\Bigr].
\end{aligned}
\]
which is exactly the claimed cumulative forgetting bound.

\section{D.  Corollary 1 Proof}
\label{coro1}
Consider a 1-dimensional \emph{torus} \(\mathbb T^1=[0,1)\) with wrap-around arithmetic, where every branch support is an interval of fixed length whose starting point is chosen uniformly in \([0,1)\). From Theorem \ref{thm:branchwise_forgetting}, taking expectation over the independent draws of the supports and using linearity, we have 
\[
\mathbb E[F_i]
\le
C
\sum_{\ell=1}^{L}
\sum_{p=1}^{d_\ell}
\sum_{q=1}^{N_\ell}
L_\ell
\sum_{j=i+1}^{T}
\mathbb E\!\Bigl[
  \mu\!\bigl(
    S^{(i)}_{\ell,p,q}\cap S^{(j)}_{\ell,p,q}
  \bigr)
\Bigr].
\tag{D-1}
\]

Fix a branch \((\ell,p,q)\). For each task index \(k\), the support \(S^{(k)}_{\ell,p,q}\subset\mathbb T^1\) is an interval
\([\,U_k,\;U_k+s_k\,)\) where
\(U_k\sim\operatorname{Uniform}[0,1)\) and all \(U_k\)’s distributions are independent. 
For a fixed point \(z\in\mathbb T^1\), 
\[
\Pr\bigl[z\in S^{(k)}_{\ell,p,q}\bigr]=s_k
\quad\text{(length of the interval).}
\]
Because the two supports for tasks \(i\) and \(j\) are independent,
\[
\Pr\bigl[
  z\in S^{(i)}_{\ell,p,q}\cap S^{(j)}_{\ell,p,q}
\bigr]
  =
  s_i\,s_j.
\]
Using \textit{Fubini’s theorem},
\[
\mathbb E\!\Bigl[
  \mu\!\bigl(
    S^{(i)}_{\ell,p,q}\cap S^{(j)}_{\ell,p,q}
  \bigr)
\Bigr]
  =
  \int_{0}^{1}
    \Pr\bigl[
      z\in S^{(i)}_{\ell,p,q}\cap S^{(j)}_{\ell,p,q}
    \bigr]dz
  =s_i\,s_j.
\tag{D-2}
\]

Next, substitute (D-2) into (D-1) and simplify. Since the expectation in (D-2) is the same for every branch, we have
\[
\begin{aligned}
\mathbb E[F_i]
&\le
C
\sum_{\ell=1}^{L}
\sum_{p=1}^{d_\ell}
\sum_{q=1}^{N_\ell}
L_\ell
\sum_{j=i+1}^{T}
s_i\,s_j
\\[4pt]
&=
C
\sum_{\ell=1}^{L}
N_\ell\,L_\ell
\sum_{j=i+1}^{T}
s_i\,s_j,
\end{aligned}
\]
which is the bound claimed in
Corollary~\ref{coro:expected_forgetting}.

\section{E.  Corollary 2 Proof}
\label{coro2}

From Theorem \ref{thm:branchwise_forgetting}, we replace the inner sum by the union measure $U^{(i)}_{\ell,p,q}$. By the sub-additivity of Lebesgue measure, we can have
\[
\mu\!\bigl(U^{(i)}_{\ell,p,q}\bigr)
  \;\le\;
  \sum_{j=i+1}^{T}
    \mu\!\bigl(
      S^{(i)}_{\ell,p,q}\!\cap S^{(j)}_{\ell,p,q}
    \bigr).
\tag{E-1}
\]
Substituting (E-1) into Theorem \ref{thm:branchwise_forgetting} yields the “saturation” bound as 
\[
F_i
\;\le\;
C
\sum_{\ell=1}^{L}
\sum_{p=1}^{d_\ell}
\sum_{q=1}^{N_\ell}
L_\ell\,
\mu\!\bigl(U^{(i)}_{\ell,p,q}\bigr),
\tag{E-2}
\]
matching the first inequality in the statement.

We then further bound the union length.
Two simple facts give an upper bound on
\(\mu(U^{(i)}_{\ell,p,q})\):

\begin{enumerate}
\item[(a)]  Each individual overlap length obeys  
            \(\mu\!\bigl(
              S^{(i)}_{\ell,p,q}\!\cap S^{(j)}_{\ell,p,q}
            \bigr)
            \le \Delta_{i,j}\);
            adding over \(j>i\) gives  
            \(\mu(U^{(i)}_{\ell,p,q})
              \le \sum_{j=i+1}^{T}\!\Delta_{i,j}\).
\item[(b)]  Since the union is contained in the branch’s own support, it guarantees that  
            \(\mu(U^{(i)}_{\ell,p,q})
              \le \mu\!\bigl(S^{(i)}_{\ell,p,q}\bigr).\)
\end{enumerate}

\noindent
Combining (a) and (b), we obtain
\[
\mu\!\bigl(U^{(i)}_{\ell,p,q}\bigr)
\le
\min\!\Bigl(
  \sum_{j=i+1}^{T}\!\Delta_{i,j},
  \,\mu\!\bigl(S^{(i)}_{\ell,p,q}\bigr)
\Bigr),
\tag{E-3}
\]
which is exactly the auxiliary bound claimed.

\medskip
\noindent
Inequalities (E-2) and (E-3) together complete the proof.

\section{F.  Theorem 3 Proof}
\label{theo3} 

Fix a radius \(r>0\) that bounds the diameter of every branch
support.  
The key geometric fact we need is that a compact
\(d_t\)-dimensional manifold can be covered by \(O(r^{-d_t})\)
\(r\)-balls.  We spell this out first, then translate it into an
overlap probability and finally into a forgetting bound.

\medskip
\noindent\emph{Covering number of \(\mathcal M_t\).}  
Endow \(\mathcal M_t\) with its intrinsic geodesic metric
\(\operatorname{dist}_{\mathcal M_t}\).
Choose a maximal set of points
\(\{x_k\}\subset\mathcal M_t\) such that the geodesic
\((r/2)\)-balls
\(\mathcal B_t(x_k,r/2)
   :=\{y\in\mathcal M_t:
       \operatorname{dist}_{\mathcal M_t}(x_k,y)\le r/2\}\)
are pairwise disjoint.
For a sufficiently small \(r\), the volume of each such ball satisfies
\[
\operatorname{Vol}_{d_t}\bigl(\mathcal B_t(x_k,r/2)\bigr)
   \;\ge\;
   c_{d_t}\,(r/2)^{d_t},
\]
where \(c_{d_t}>0\) depends only on dimension (it is the Euclidean
unit-ball volume, up to a curvature factor bounded away from zero in
the compact domain).
Because the disjoint balls lie inside \(\mathcal M_t\), we have
\[
\#\{x_k\}
  \;\le\;
  \frac{\operatorname{Vol}_{d_t}(\mathcal M_t)}
       {c_{d_t}(r/2)^{d_t}}
  \;=\;
  O\!\bigl(r^{-d_t}\bigr).
\]
Maximality implies that the concentric \(r\)-balls
\(\mathcal B_t(x_k,r)\) cover \(\mathcal M_t\); hence the covering
number obeys
\[
N_t(r)=O\!\bigl(r^{-d_t}\bigr).
\]

\medskip
\noindent\emph{Probability that a branch fires on task \(t\).}  
For each branch \((\ell,p,q)\) the support on task \(t\) is assumed to
sit inside one of the \(N_t(r)\) covering balls, chosen uniformly at
random.  A fixed pre-activation coordinate \(z\) is therefore contained
in that support with probability
\[
\Pr\!\bigl\{z\in S^{(t)}_{\ell,p,q}\bigr\}
   =\frac{1}{N_t(r)}
   =O\!\bigl(r^{d_t}\bigr).
\]

\medskip
\noindent\emph{Expected overlap for tasks \(i\) and \(j\).}  
Independent sampling for the two tasks gives
\[
\mathbb E\!\bigl[
  \mu\!\bigl(
    S^{(i)}_{\ell,p,q}\cap S^{(j)}_{\ell,p,q}
  \bigr)
\bigr]
 =\! \int_{0}^{1}\!
    O\!\bigl(r^{d_i}\bigr)\,
    O\!\bigl(r^{d_j}\bigr)\,dz
 =O\!\bigl(r^{d_i+d_j}\bigr).
\tag{F-1}
\]

\noindent
Insert the overlap into the cumulative bound.  
Taking expectation on the branch-wise cumulative inequality in Theorem~\ref{thm:branchwise_forgetting} and substituting (F-1), we obtain
\[
\mathbb E[F_i]
  \le
  C\Bigl(\sum_{\ell}N_\ell L_\ell\Bigr)
  r^{d_i}\!
  \sum_{j=i+1}^{T} r^{d_j}.
\tag{F-2}
\]
With \(N_{\text{tot}}:=\sum_\ell N_\ell\) and
\(\bar L:=N_{\text{tot}}^{-1}\sum_\ell N_\ell L_\ell\), 
equation (F-2) becomes
\[
F_i
 =O\!\Bigl(
   N_{\text{tot}}\,
   \bar L
   \sum_{j=i+1}^{T} r^{\,d_i+d_j}
 \Bigr),
\]
which is the intrinsic-dimension forgetting rate stated.

\section{G.  Corollary 3 Proof}
\label{coro3} 

In Theorem~\ref{thm:intrinsic_dimension}, with \(d_j\le D\) for every \(j>i\), each term in the sum is at most
\(r^{\,d_i+D}\). Since there are \(T-i\le T\) such terms, we thus have  
\[
\sum_{j=i+1}^{T} r^{\,d_i+d_j}
  \;\le\;
  T\,r^{\,d_i+D}.
\tag{G-1}
\]
Substituting (G-1) into Theorem~\ref{thm:intrinsic_dimension} yields  
\[
F_i
  =O\!\bigl(
      T\,N_{\mathrm{tot}}\,
      \bar L\;
      r^{\,d_i+D}
    \bigr),
\]
proving the corollary.  
Because \(d_i+D\) is sufficiently small, the factor \(r^{\,d_i+D}\) can
drive the bound arbitrarily close to a constant rate, demonstrating
the robust retention performance for low-dimensional tasks.

\section{H.  Corollary 4 Proof}
\label{coro4} 

Assume that, for every task \(t\), the support of each branch is
partitioned into \(k_t\) disjoint intervals of radius
\(r/k_t\). Repeating the covering argument with this reduced radius replaces each factor \(r^{d_t}\) in Theorem~\ref{thm:intrinsic_dimension} by \(\bigl(r/k_t\bigr)^{d_t}\).
Applying the substitution for tasks \(i\) and \(j\) gives
\[
F_i
  =O\!\Bigl(
     N_{\mathrm{tot}}\,
     \bar L
     \sum_{j=i+1}^{T}
       \bigl(r/k_i\bigr)^{d_i}\!
       \bigl(r/k_j\bigr)^{d_j}
   \Bigr), 
\]
which matches the improved forgetting rate. Consequently, forgetting decays as
\(\,(k_i k_j)^{-(d_i+d_j)}\), indicating that splitting every branch of task \(t\) into \(k_t\) pieces diminishes that task’s overlap contribution by the factor \(k_t^{-d_t}\).

\section{I.  Binary and Decimal Addition}

This section provides further details of our experiments on binary and decimal addition, including the KAN model architectures and the benchmark/baseline used for evaluation. 

\medskip

\noindent
\textbf{1. Additional Setups.} 
\medskip

\noindent
In both the binary and decimal addition experiments, we define a sequence of five tasks. Task 1, termed one’s addition, involves adding the digit 1 to each digit from 1 to 9 (e.g., 1 + 1 = 2, through 1 + 9 = 10, and 1 + 1 = 2, through 9 + 1 = 10). The subsequent tasks, \textit{two’s}, \textit{three’s}, \textit{four’s}, and \textit{five’s}, are similarly constructed by adding 2, 3, 4, and 5, respectively, to each digit in the same range. In the binary addition experiments, all input digits are represented as 4-bit binary numbers. Tables~\ref{tab:binary_addition_tasks} and~\ref{tab:decimal_addition_tasks} summarize the synthetic datasets used for the binary and decimal addition tasks, respectively. 

\begin{table}[t]
\centering
\resizebox{\columnwidth}{!}{%
\begin{tabular}{c|c|c|c|c}
\toprule
\textbf{Task 1} & \textbf{Task 2} & \textbf{Task 3} & \textbf{Task 4} & \textbf{Task 5} \\
\cmidrule{1-5}
0001 + 0001 & 0010 + 0001 & 0011 + 0001 & 0100 + 0001 & 0101 + 0001 \\
0001 + 0010 & 0010 + 0010 & 0011 + 0010 & 0100 + 0010 & 0101 + 0010 \\
0001 + 0011 & 0010 + 0011 & 0011 + 0011 & 0100 + 0011 & 0101 + 0011 \\
0001 + 0100 & 0010 + 0100 & 0011 + 0100 & 0100 + 0100 & 0101 + 0100 \\
0001 + 0101 & 0010 + 0101 & 0011 + 0101 & 0100 + 0101 & 0101 + 0101 \\
0001 + 0110 & 0010 + 0110 & 0011 + 0110 & 0100 + 0110 & 0101 + 0110 \\
0001 + 0111 & 0010 + 0111 & 0011 + 0111 & 0100 + 0111 & 0101 + 0111 \\
0001 + 1000 & 0010 + 1000 & 0011 + 1000 & 0100 + 1000 & 0101 + 1000 \\
0001 + 1001 & 0010 + 1001 & 0011 + 1001 & 0100 + 1001 & 0101 + 1001 \\
0001 + 0001 & 0001 + 0010 & 0001 + 0011 & 0001 + 0100 & 0001 + 0101 \\
0010 + 0001 & 0010 + 0010 & 0010 + 0011 & 0010 + 0100 & 0010 + 0101 \\
0011 + 0001 & 0011 + 0010 & 0011 + 0011 & 0011 + 0100 & 0011 + 0101 \\
0100 + 0001 & 0100 + 0010 & 0100 + 0011 & 0100 + 0100 & 0100 + 0101 \\
0101 + 0001 & 0101 + 0010 & 0101 + 0011 & 0101 + 0100 & 0101 + 0101 \\
0110 + 0001 & 0110 + 0010 & 0110 + 0011 & 0110 + 0100 & 0110 + 0101 \\
0111 + 0001 & 0111 + 0010 & 0111 + 0011 & 0111 + 0100 & 0111 + 0101 \\
1000 + 0001 & 1000 + 0010 & 1000 + 0011 & 1000 + 0100 & 1000 + 0101 \\
1001 + 0001 & 1001 + 0010 & 1001 + 0011 & 1001 + 0100 & 1001 + 0101 \\
\bottomrule
\end{tabular}
}
\caption{Binary addition tasks.}
\label{tab:binary_addition_tasks}
\end{table}

\begin{table}[t]
\centering
\resizebox{0.65\columnwidth}{!}{%
\begin{tabular}{c|c|c|c|c}
\toprule
\textbf{Task 1} & \textbf{Task 2} & \textbf{Task 3} & \textbf{Task 4} & \textbf{Task 5} \\
\midrule
1 + 1 & 2 + 1 & 3 + 1 & 4 + 1 & 5 + 1 \\
1 + 2 & 2 + 2 & 3 + 2 & 4 + 2 & 5 + 2 \\
1 + 3 & 2 + 3 & 3 + 3 & 4 + 3 & 5 + 3 \\
1 + 4 & 2 + 4 & 3 + 4 & 4 + 4 & 5 + 4 \\
1 + 5 & 2 + 5 & 3 + 5 & 4 + 5 & 5 + 5 \\
1 + 6 & 2 + 6 & 3 + 6 & 4 + 6 & 5 + 6 \\
1 + 7 & 2 + 7 & 3 + 7 & 4 + 7 & 5 + 7 \\
1 + 8 & 2 + 8 & 3 + 8 & 4 + 8 & 5 + 8 \\
1 + 9 & 2 + 9 & 3 + 9 & 4 + 9 & 5 + 9 \\
1 + 1 & 1 + 2 & 1 + 3 & 1 + 4 & 1 + 5 \\
2 + 1 & 2 + 2 & 2 + 3 & 2 + 4 & 2 + 5 \\
3 + 1 & 3 + 2 & 3 + 3 & 3 + 4 & 3 + 5 \\
4 + 1 & 4 + 2 & 4 + 3 & 4 + 4 & 4 + 5 \\
5 + 1 & 5 + 2 & 5 + 3 & 5 + 4 & 5 + 5 \\
6 + 1 & 6 + 2 & 6 + 3 & 6 + 4 & 6 + 5 \\
7 + 1 & 7 + 2 & 7 + 3 & 7 + 4 & 7 + 5 \\
8 + 1 & 8 + 2 & 8 + 3 & 8 + 4 & 8 + 5 \\
9 + 1 & 9 + 2 & 9 + 3 & 9 + 4 & 9 + 5 \\
\bottomrule
\end{tabular}
}
\caption{Decimal addition tasks.}
\label{tab:decimal_addition_tasks}
\end{table}

\begin{table}[h]
\centering
\resizebox{0.7\columnwidth}{!}{%
\begin{tabular}{l|l|l}
\toprule
\textbf{Category} & \textbf{Hyperparameter} & \textbf{Value} \\
\midrule
\multirow{6}{*}{Model Architecture} 
& Hidden Layers & [3, 2, 2] \\
& Grid Size  & 5 \\
& Spline Order  & 3 \\
& Base Activation & SiLU \\
& Grid Range & [-1, 1] \\
& Grid Epsilon  & 0.02 \\
\midrule
\multirow{4}{*}{Initialization / Scaling}
& Base Weight Scale & 1.0 \\
& Spline Weight Scale & 1.0 \\
& Spline Noise Scale & 0.1 \\
& Enable Spline Scaler & True \\
\midrule
\multirow{3}{*}{Optimizer \& Training}
& Optimizer & AdamW \\
& Learning Rate & 1e-3 \\
& Weight Decay & 1e-4 \\
\midrule
\multirow{2}{*}{Loss}
& Loss Function & MSE Loss \\
& Prediction Threshold & 0.5 \\
\midrule
\multirow{1}{*}{Training Loop}
& Epochs per Task & 50 \\

\bottomrule
\end{tabular}
}
\caption{Hyperparameters of the KAN in binary addition.}
\label{tab:binary_kan_hyperparams}
\end{table}

\begin{table}[h]
\centering
\resizebox{\columnwidth}{!}{%
\begin{tabular}{l|l|l}
\toprule
\textbf{Category} & \textbf{Hyperparameter} & \textbf{Value} \\
\midrule
\multirow{5}{*}{Model Architecture} 
& Hidden Layer Dimensions & [2, 3, 2] \\
& Grid Size  & 5 \\
& Spline Order & 3 \\
& Base Activation & SiLU \\
& Grid Range & [-1, 1] \\
\midrule
\multirow{4}{*}{Initialization / Scaling}
& Base Weight Scale & 1.0 \\
& Spline Weight Scale & 1.0 \\
& Noise Scale & 0.1 \\
& Enable Spline Scaler & True \\
\midrule
\multirow{4}{*}{Training}
& Optimizer & AdamW \\
& Learning Rate & 1e-3 \\
& Weight Decay & 1e-4 \\
& Epochs per Task & 100 \\
\midrule
\multirow{2}{*}{Loss and Evaluation}
& Loss Function & MSE Loss \\
& Output Format & \texttt{[Sum (mod 10), Carry Bit]} \\
\midrule
\multirow{1}{*}{Support Overlap}
& Activation Threshold ($t$) & 1e-2 \\
\bottomrule
\end{tabular}
}
\caption{Hyperparameters of the KAN in decimal addition.}
\label{tab:decimal_kan_hyperparams}
\end{table}

\begin{figure}[h]
    \centering

    \begin{subfigure}[b]{0.09\textwidth}
        \includegraphics[width=\textwidth]{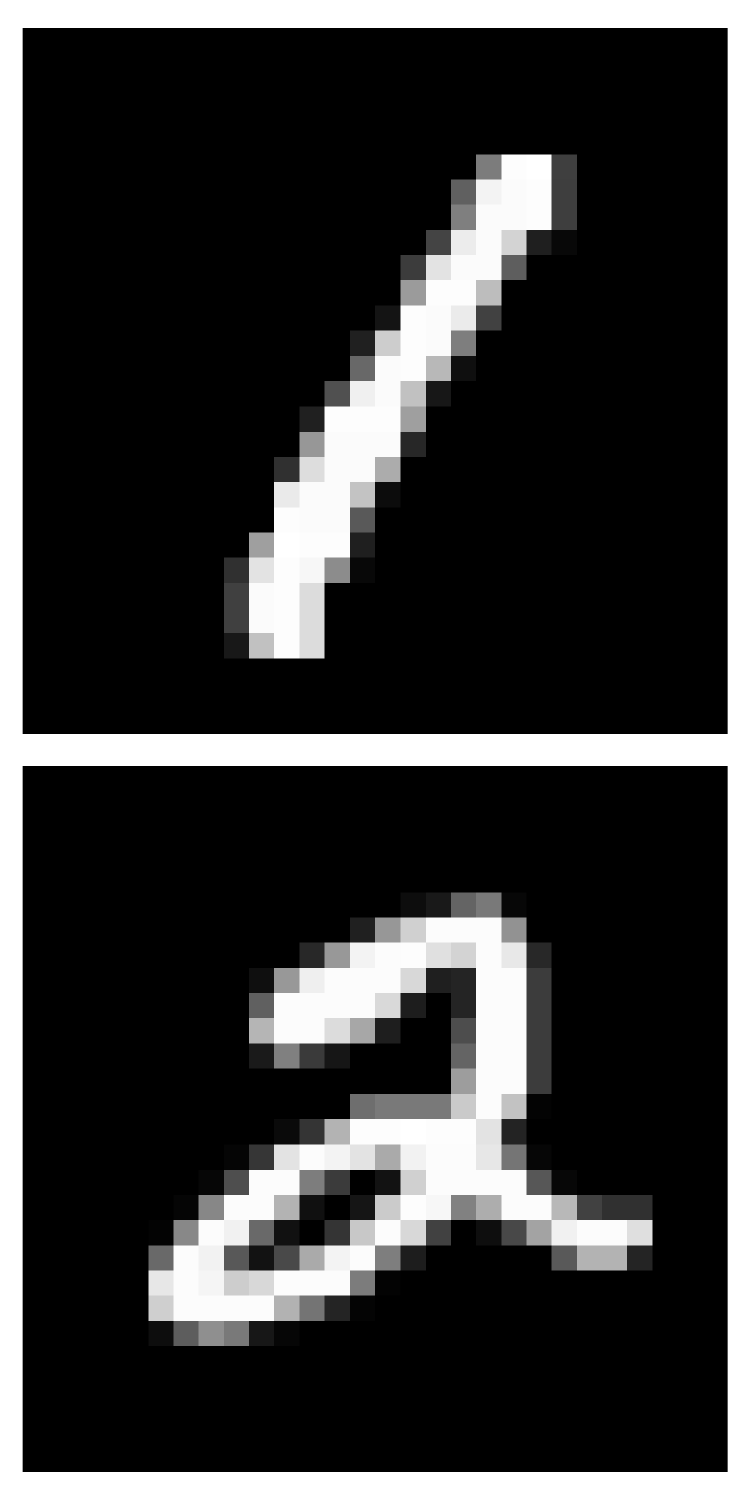}
        \caption{}
        \label{fig:mn1}
    \end{subfigure}
    \begin{subfigure}[b]{0.09\textwidth}
        \includegraphics[width=\textwidth]{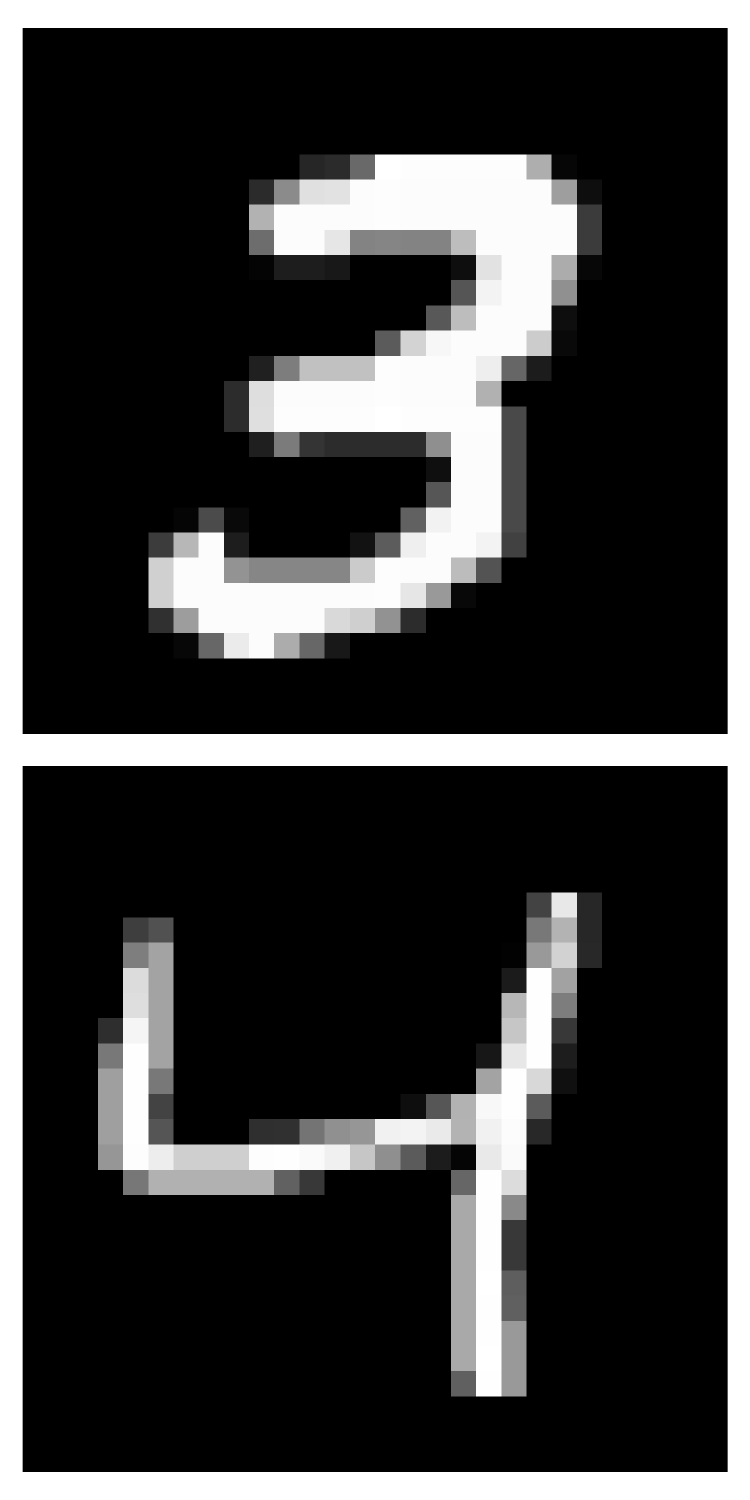}
        \caption{}
        \label{fig:mn2}
    \end{subfigure}
    \begin{subfigure}[b]{0.09\textwidth}
        \includegraphics[width=\textwidth]{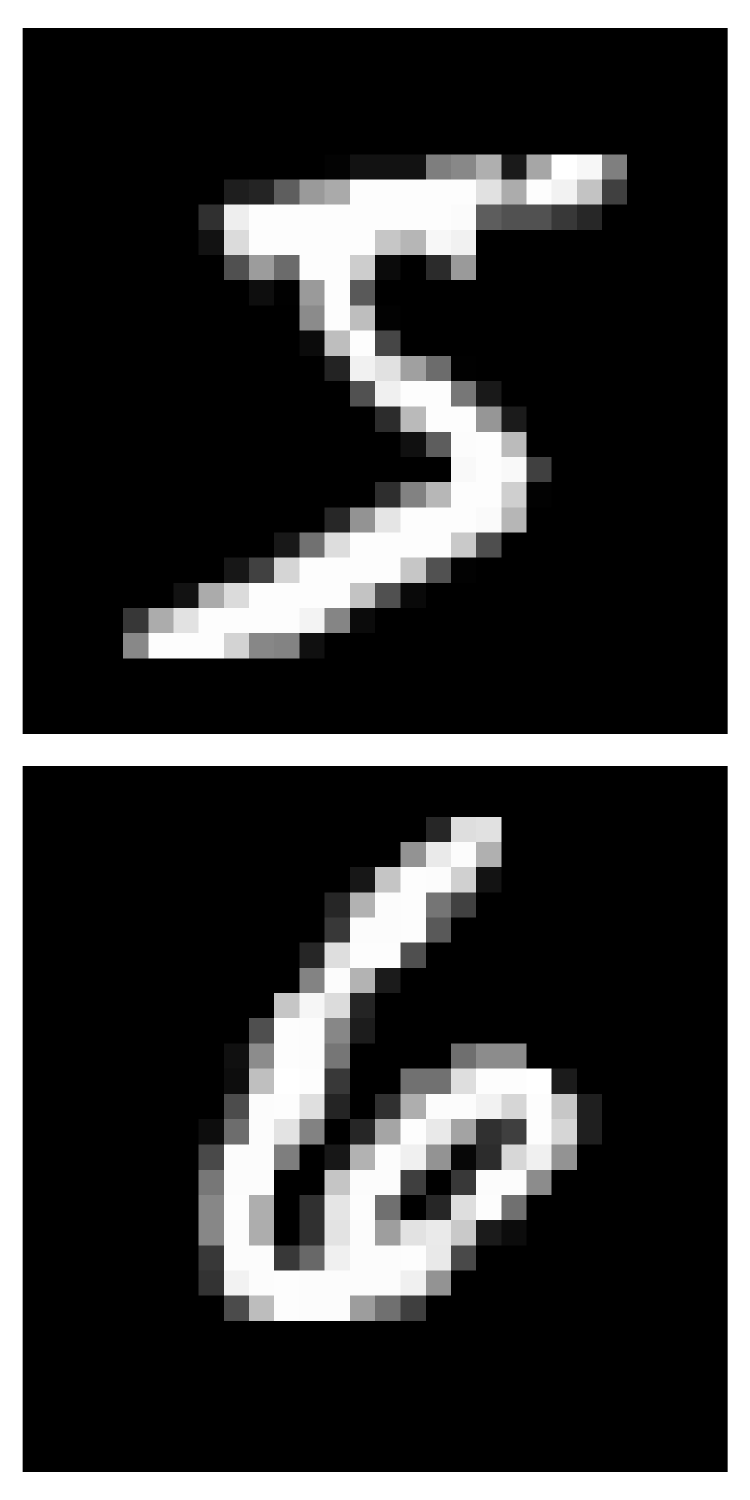}
        \caption{}
        \label{fig:mn3}
    \end{subfigure}
    \begin{subfigure}[b]{0.09\textwidth}
        \includegraphics[width=\textwidth]{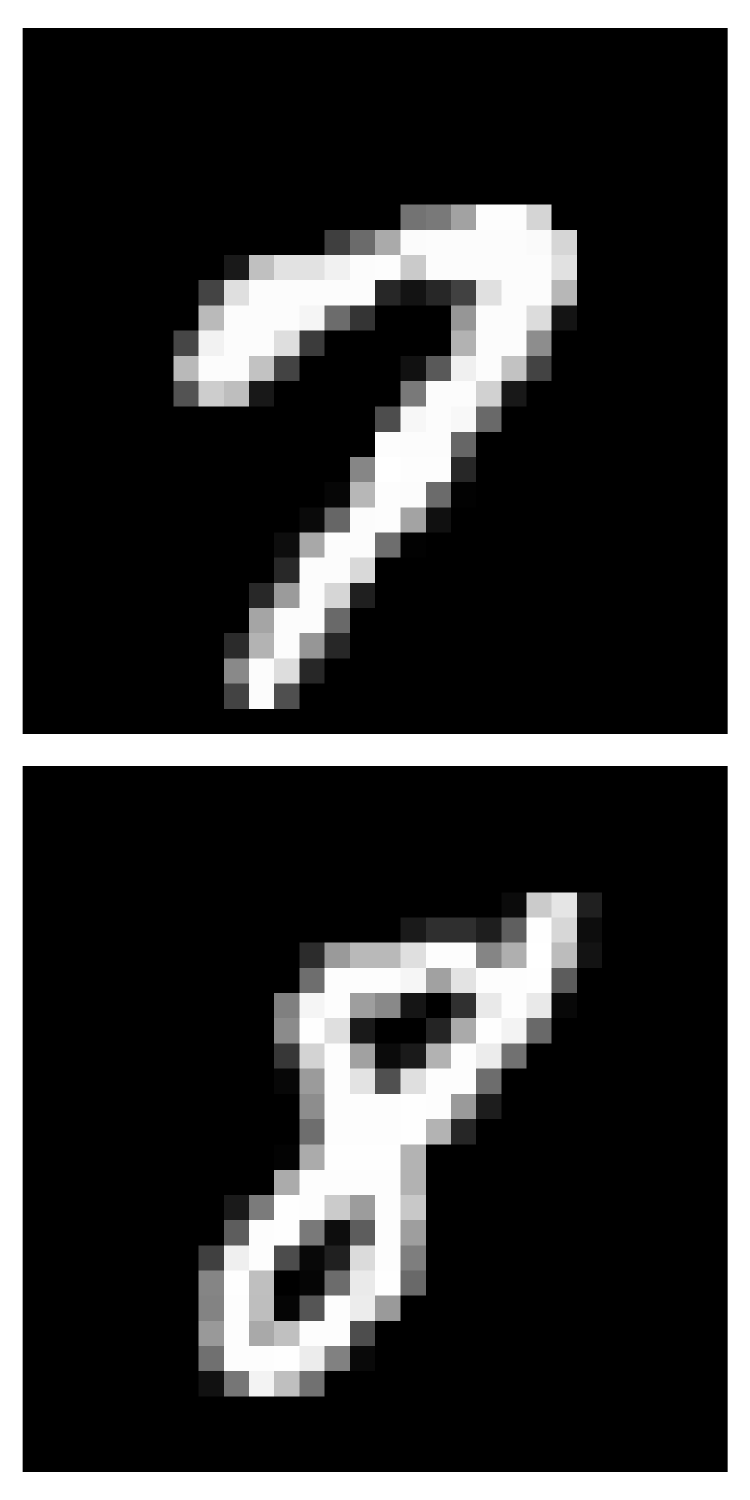}
        \caption{}
        \label{fig:mn4}
    \end{subfigure}
    \begin{subfigure}[b]{0.09\textwidth}
        \includegraphics[width=\textwidth]{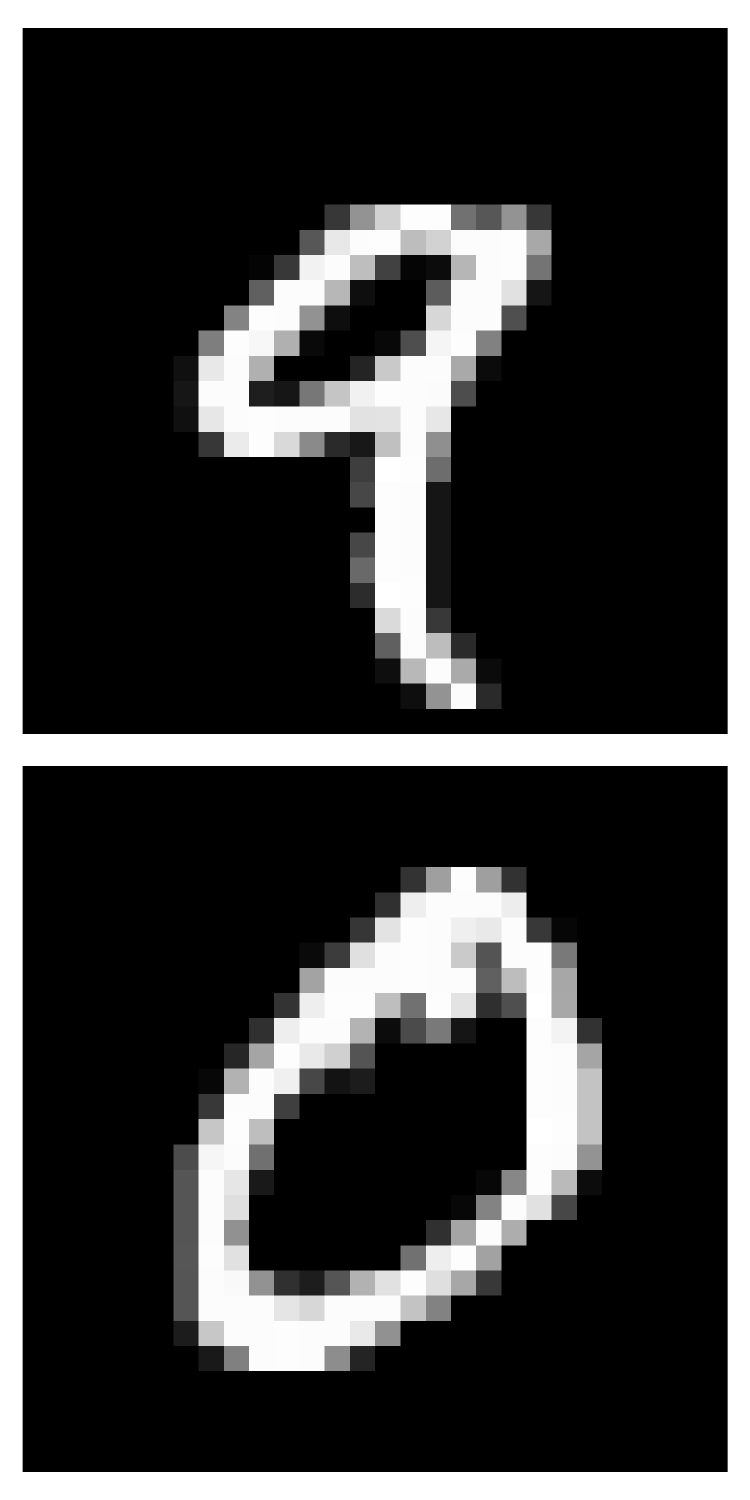}
        \caption{}
        \label{fig:mn5}
    \end{subfigure}
    \begin{subfigure}[b]{0.09\textwidth}
        \includegraphics[width=\textwidth]{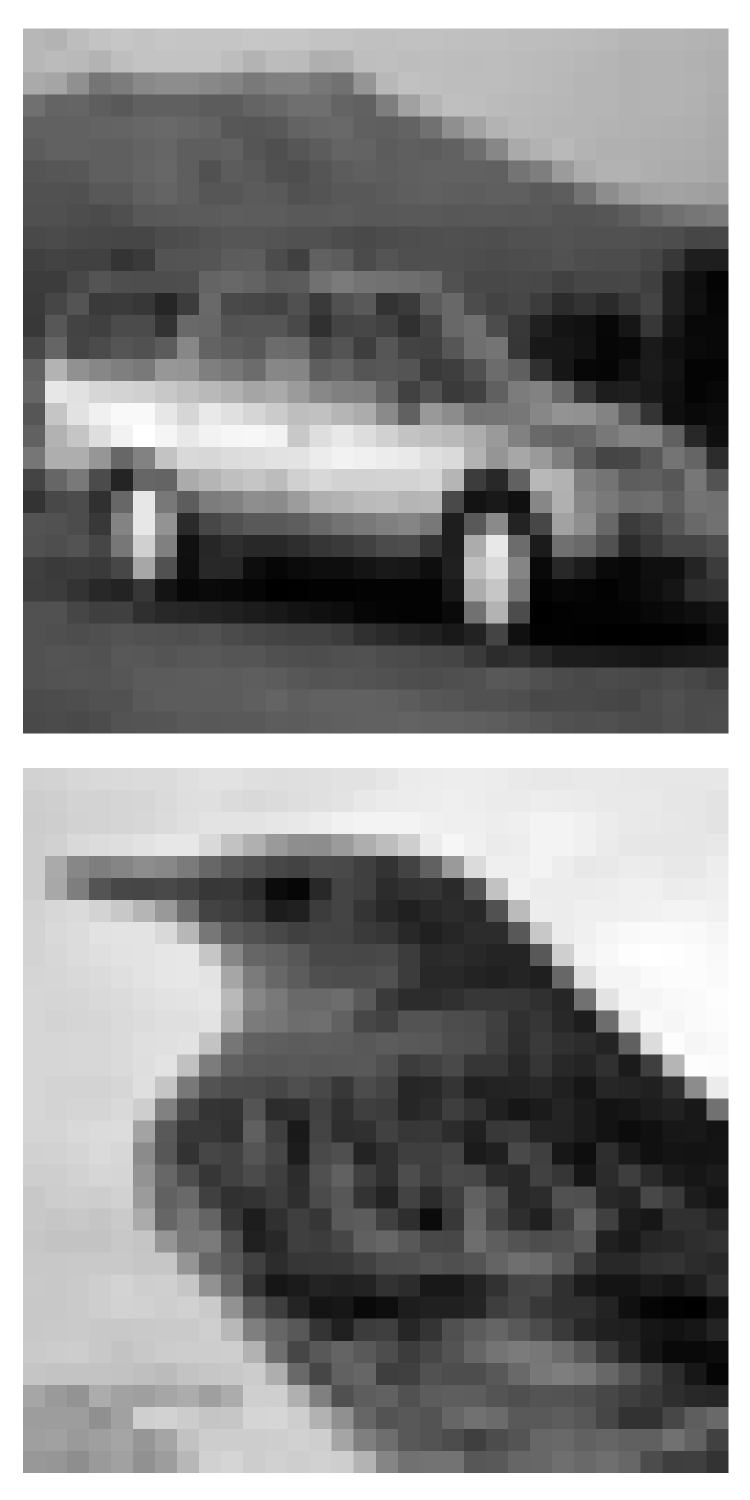}
        \caption{}
        \label{fig:ct1}
    \end{subfigure}
    \begin{subfigure}[b]{0.09\textwidth}
        \includegraphics[width=\textwidth]{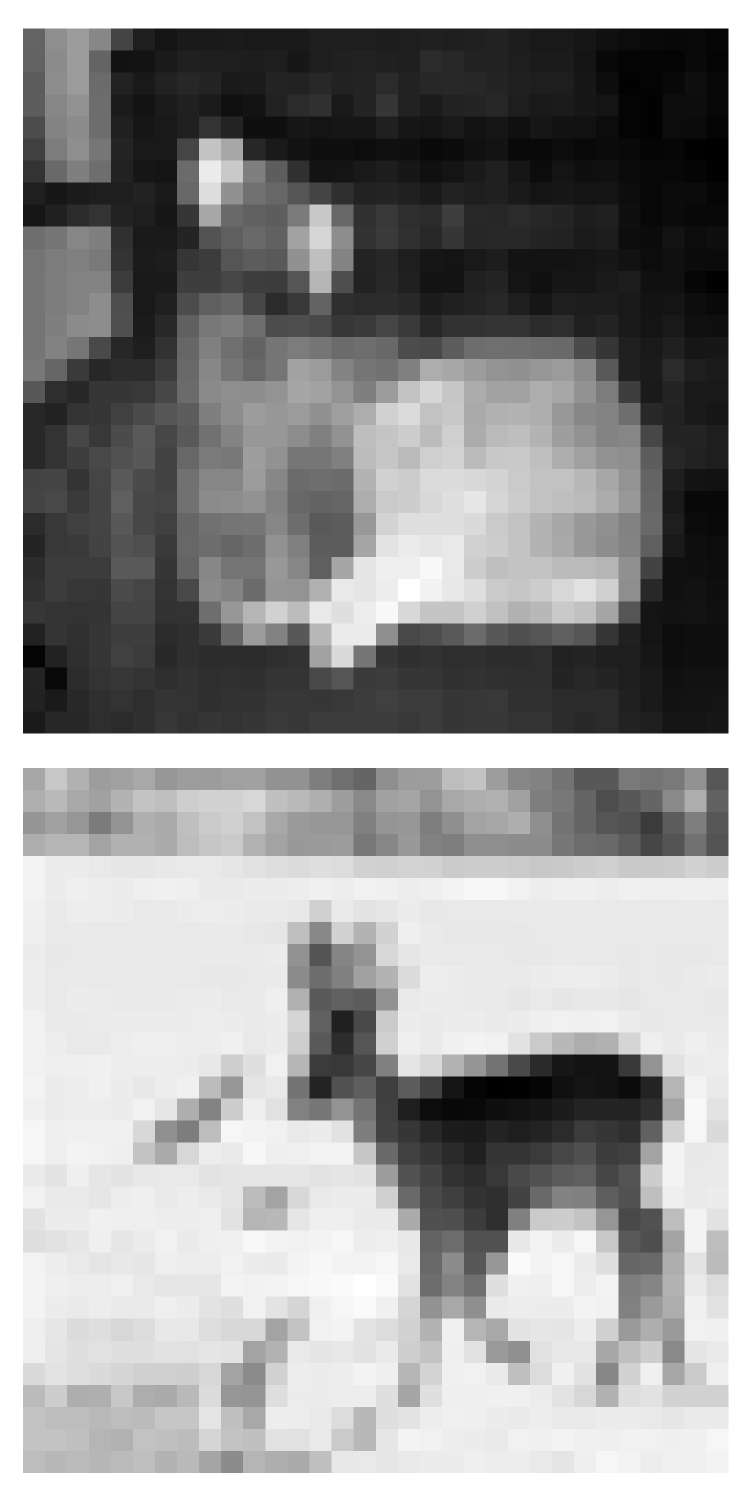}
        \caption{}
        \label{fig:ct2}
    \end{subfigure}
    \begin{subfigure}[b]{0.09\textwidth}
        \includegraphics[width=\textwidth]{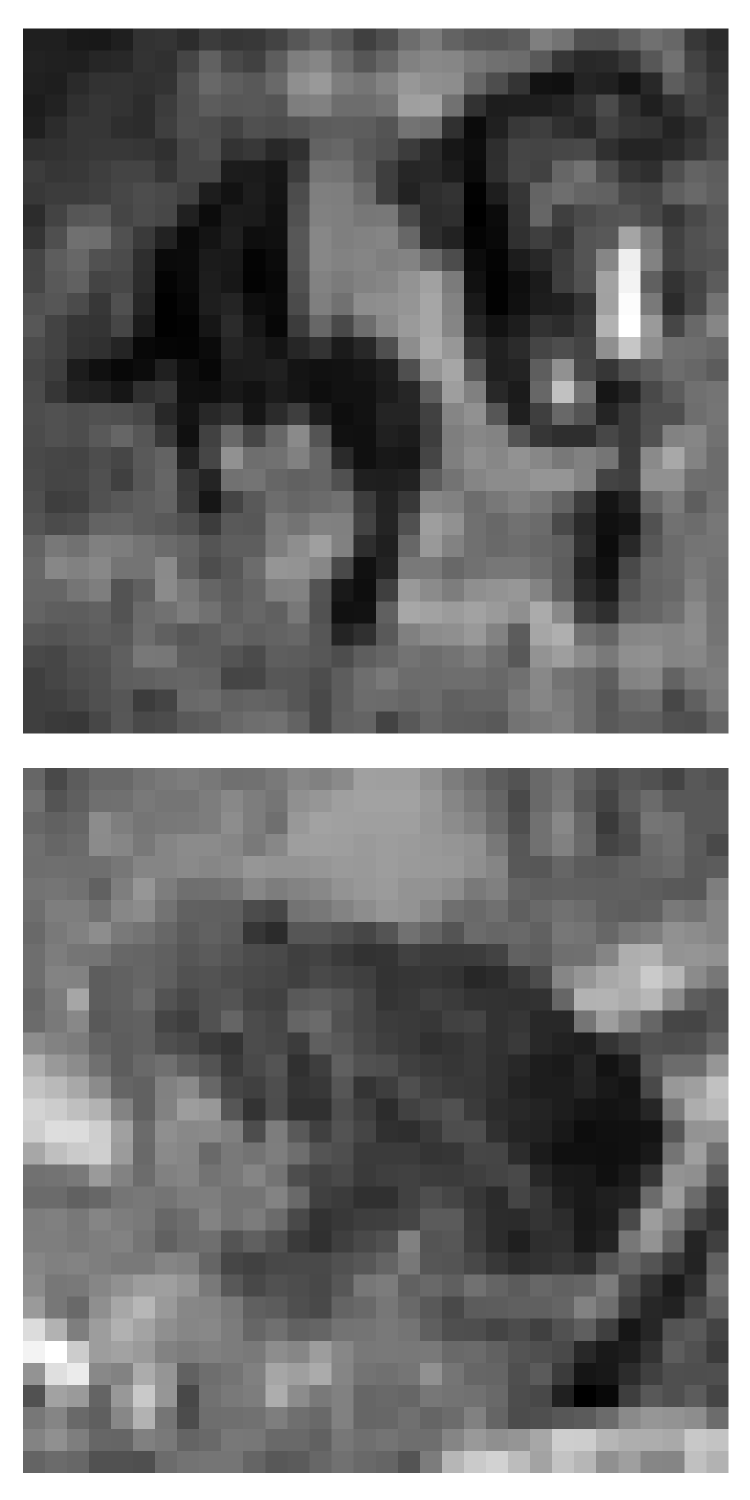}
        \caption{}
        \label{fig:ct3}
    \end{subfigure}
    \begin{subfigure}[b]{0.09\textwidth}
        \includegraphics[width=\textwidth]{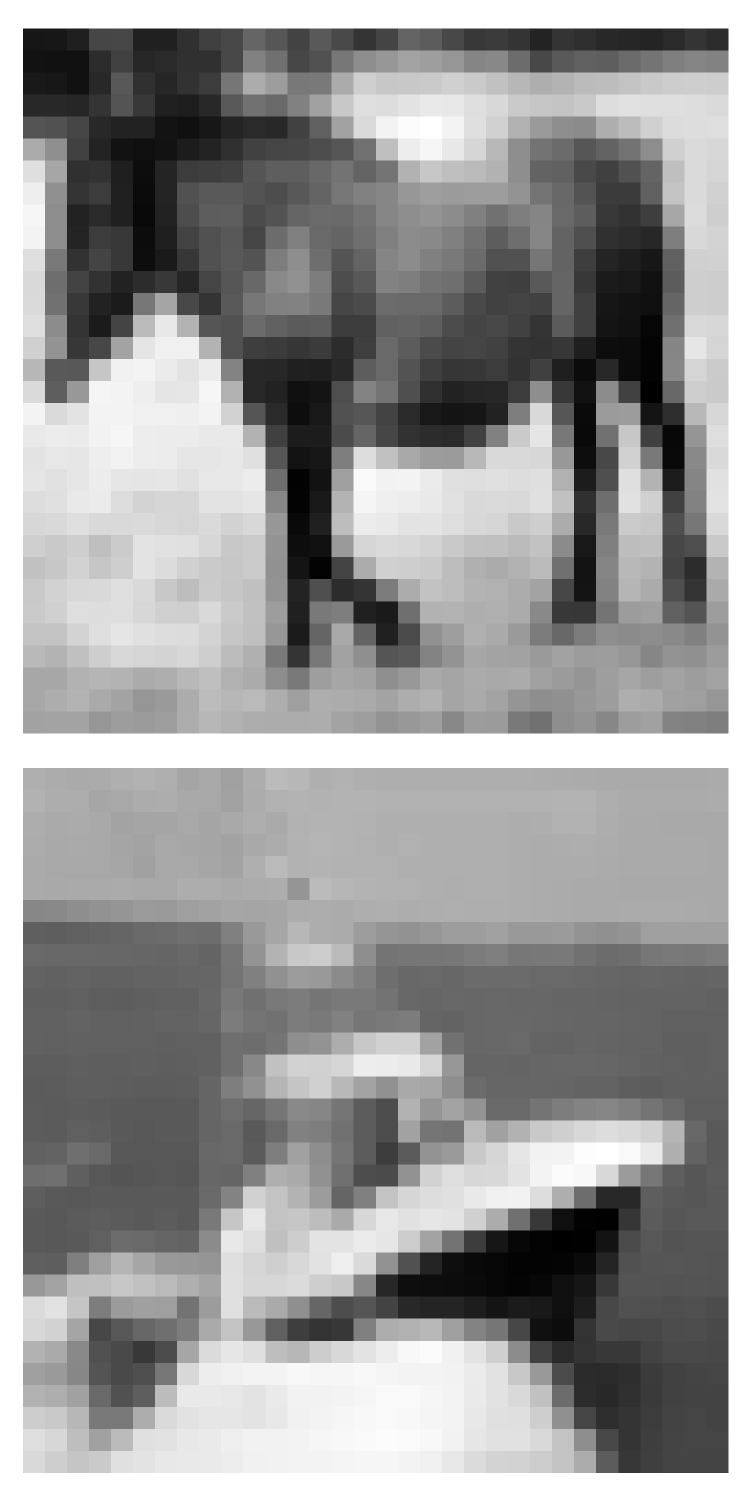}
        \caption{}
        \label{fig:ct4}
    \end{subfigure}
    \begin{subfigure}[b]{0.09\textwidth}
        \includegraphics[width=\textwidth]{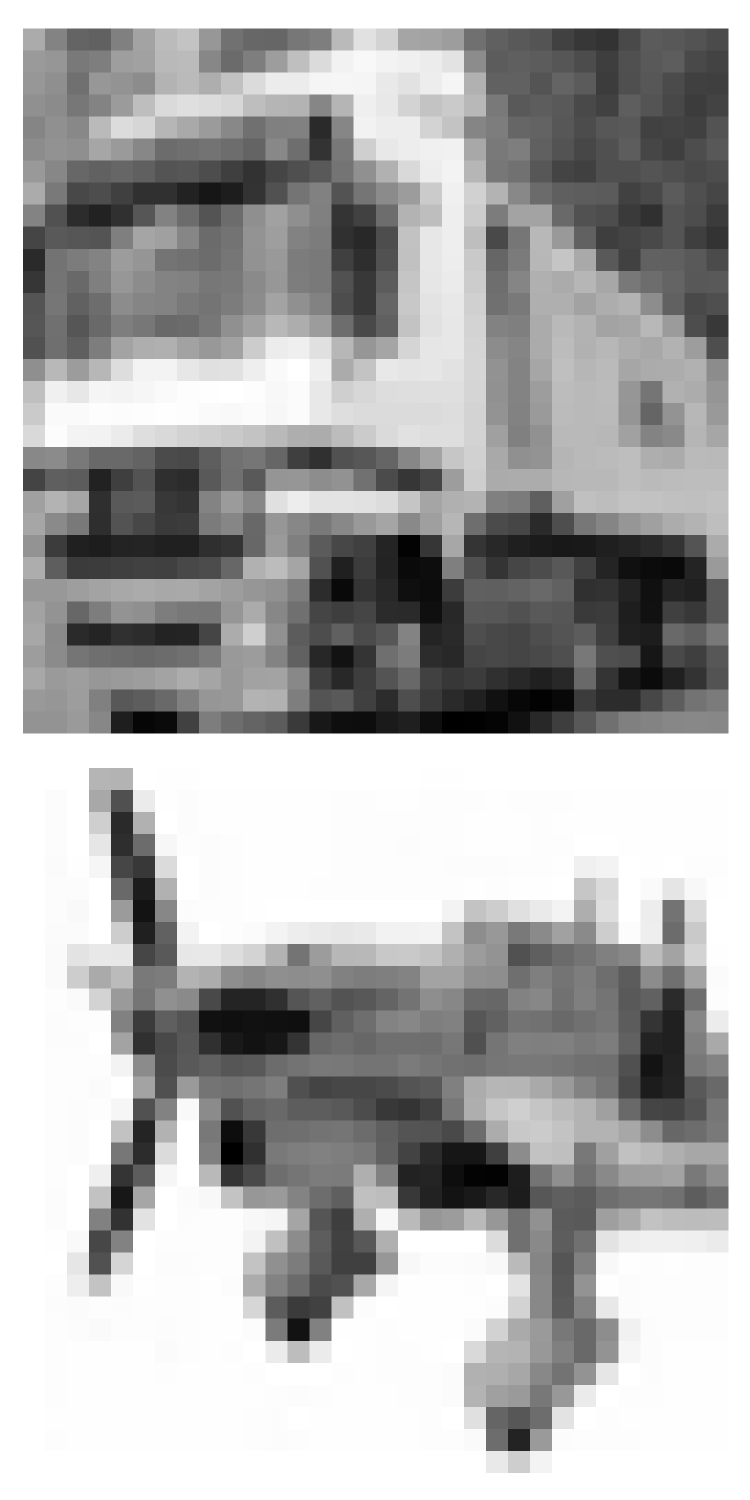}
        \caption{}
        \label{fig:ct5}
    \end{subfigure}
    \begin{subfigure}[b]{0.09\textwidth}
        \includegraphics[width=\textwidth]{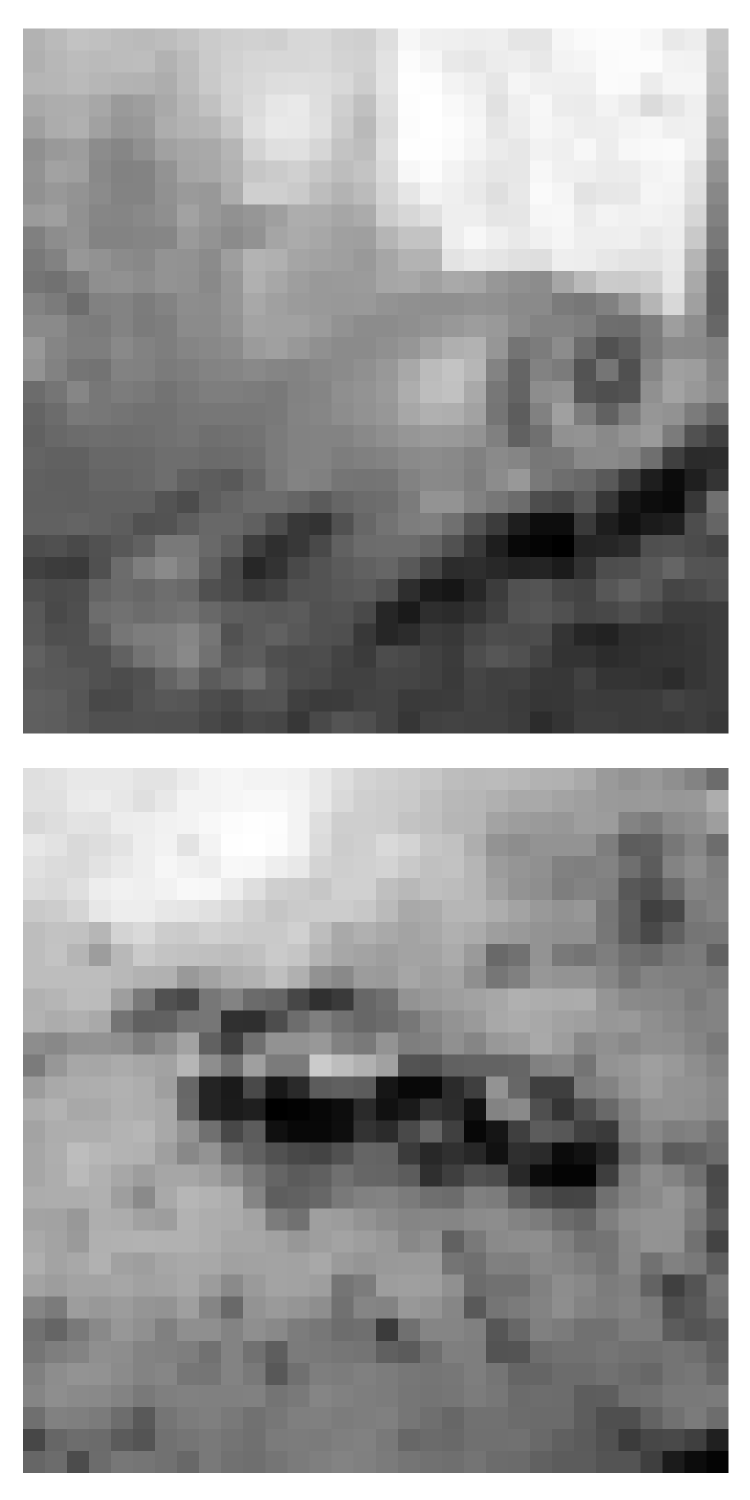}
        \caption{}
        \label{fig:it1}
    \end{subfigure}
    \begin{subfigure}[b]{0.09\textwidth}
        \includegraphics[width=\textwidth]{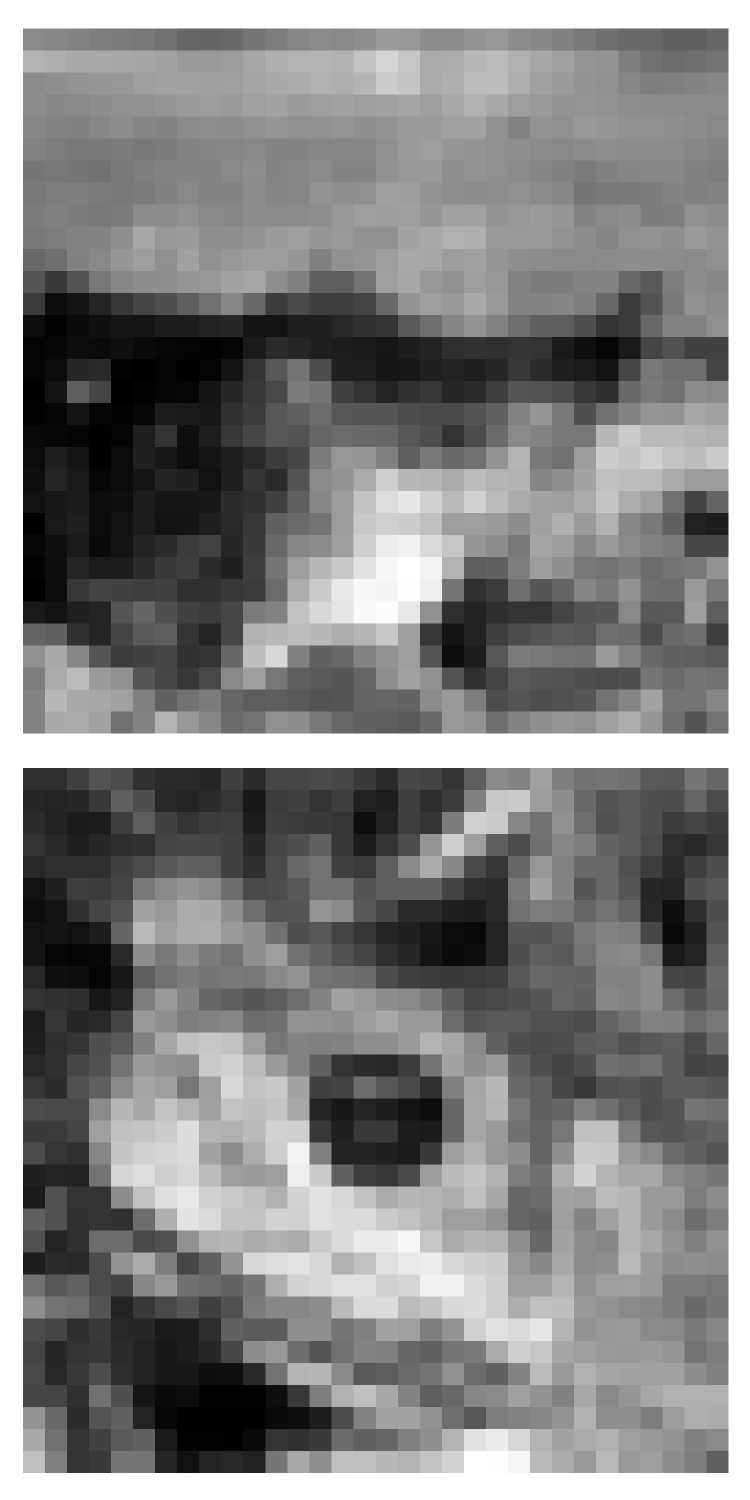}
        \caption{}
        \label{fig:it2}
    \end{subfigure}
    \begin{subfigure}[b]{0.09\textwidth}
        \includegraphics[width=\textwidth]{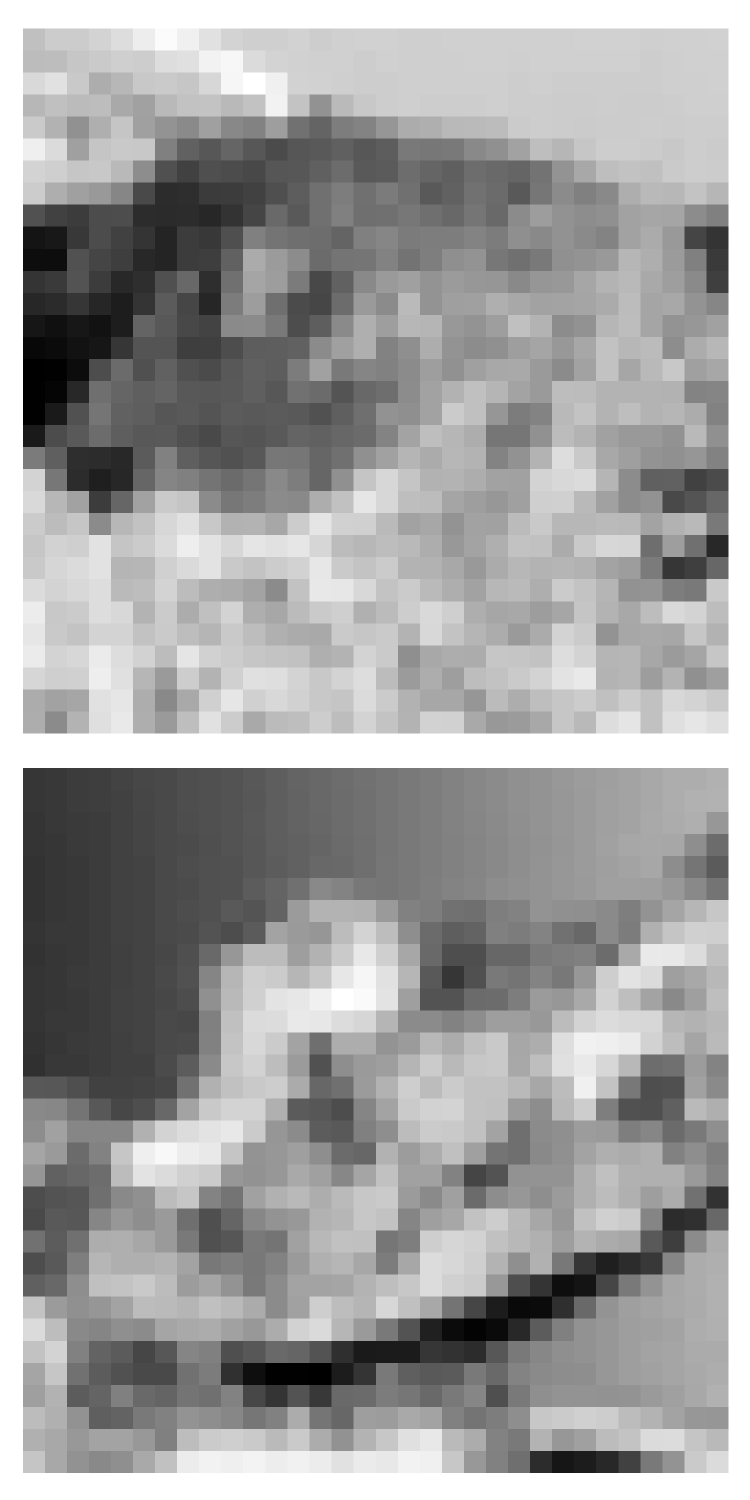}
        \caption{}
        \label{fig:it3}
    \end{subfigure}
    \begin{subfigure}[b]{0.09\textwidth}
        \includegraphics[width=\textwidth]{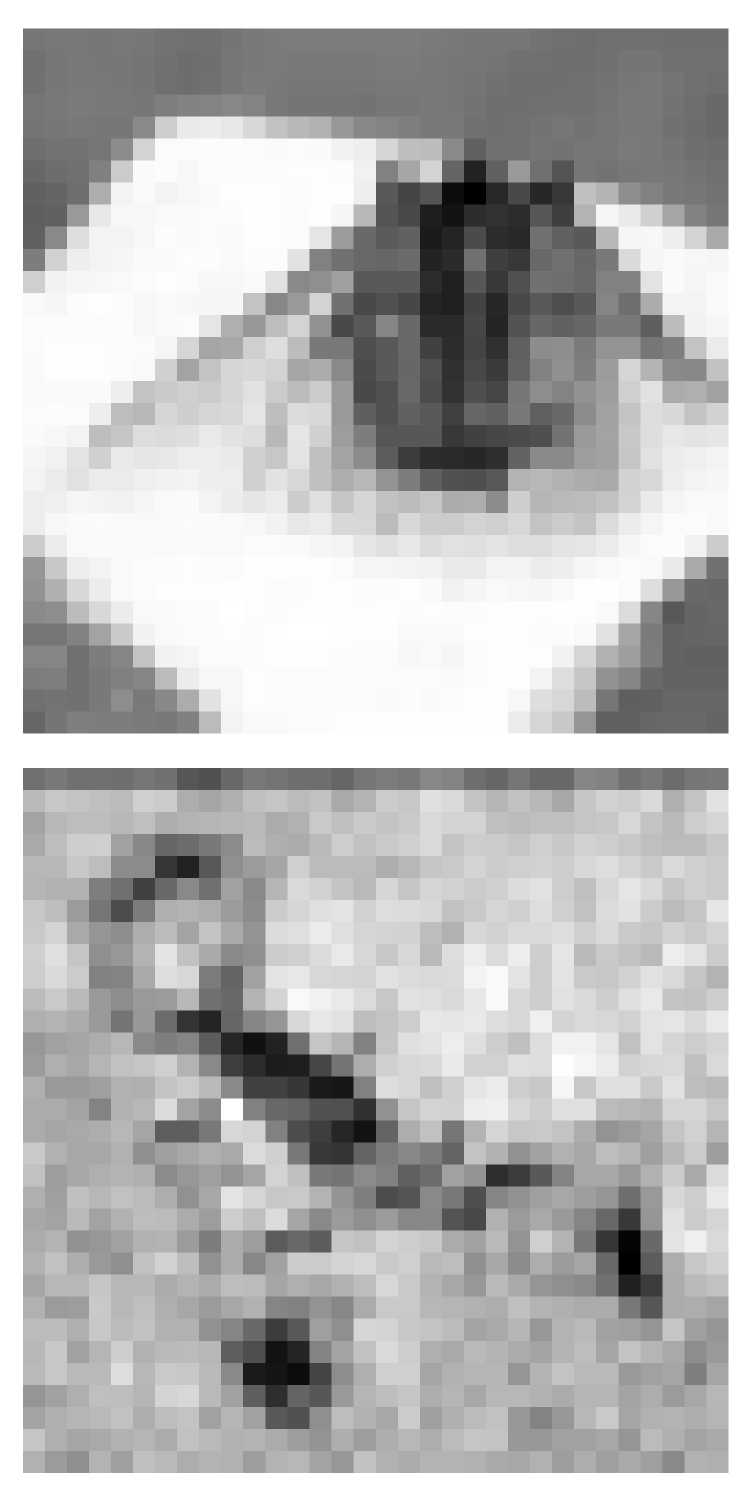}
        \caption{}
        \label{fig:it4}
    \end{subfigure}
    \begin{subfigure}[b]{0.09\textwidth}
        \includegraphics[width=\textwidth]{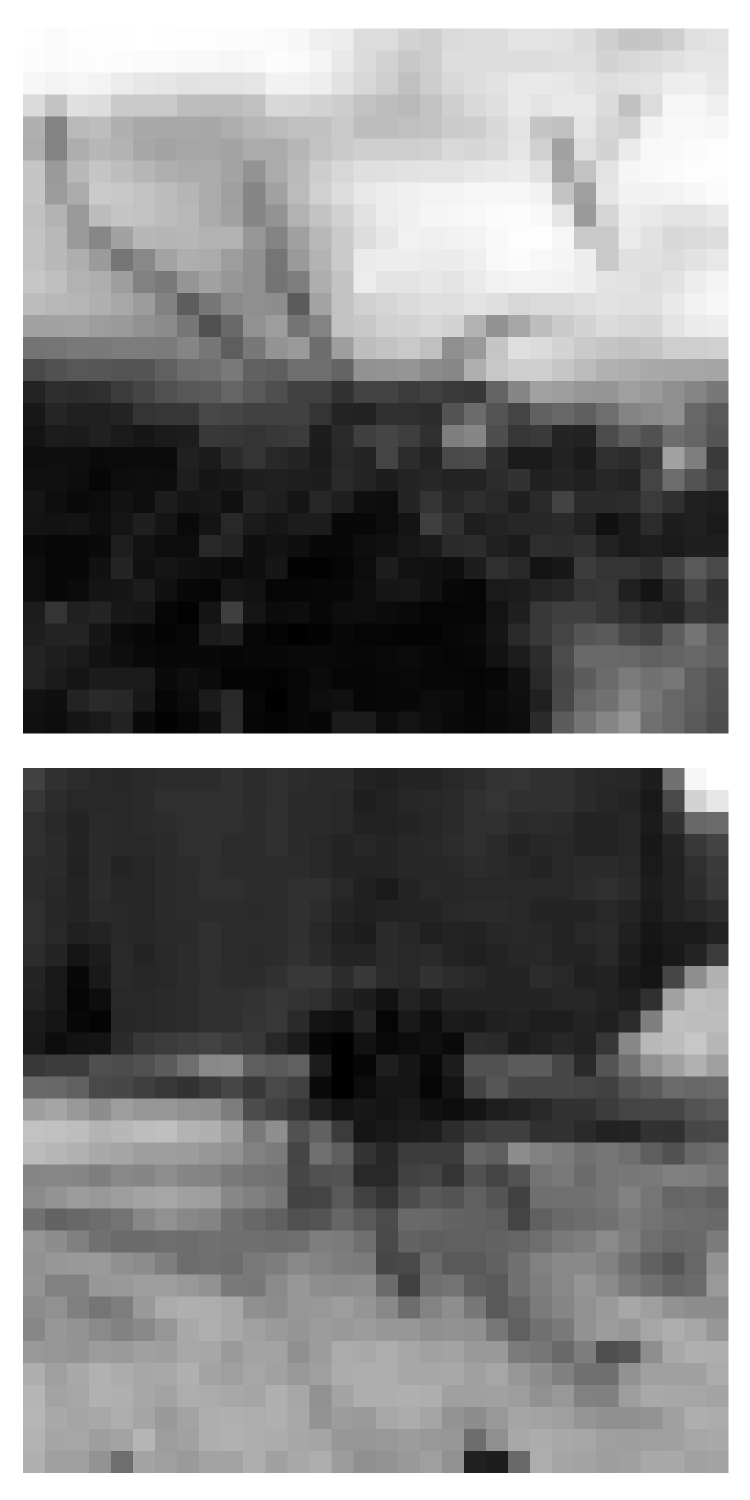}
        \caption{}
        \label{fig:it5}
    \end{subfigure}
    \caption{Five sequential tasks were respectively constructed for MNIST~(\subref{fig:mn1}-\subref{fig:mn5}), CIFAR-10~(\subref{fig:ct1}-\subref{fig:ct5}) and Tiny-ImageNet~(\subref{fig:it1}-\subref{fig:it5}) datasets, where each task containing two distinct classes.} 
    \label{fig:cifar}
\end{figure}

\medskip
\noindent
\textbf{2. Additional Configurations.} 
\medskip

\noindent
The KAN model used for the binary addition tasks takes three inputs at each time step, one bit from each of the two 4-bit binary numbers, and a carry-in bit from the previous step. The initial carry-in bit for the least significant bit addition is set to zero. These inputs are passed into the KAN, which outputs the current sum bit and a carry-out to be used in the next step. This carry-out is recurrently fed back into the model, enabling sequential processing of the bit pairs from the least to the most significant positions. In contrast, the KAN model used for decimal addition takes two decimal digits as input and produces the corresponding sum digit and carry-out. Tables~\ref{tab:binary_kan_hyperparams} and~\ref{tab:decimal_kan_hyperparams} show the hyperparameters for the model architectures used in both synthetic tasks.

\medskip
\noindent
The baseline~\cite{binaryaddition} used for comparison with the KAN on continual binary addition tasks is based on the hypothesis that, since addition is an algorithmic process, a suitably designed model architecture can avoid catastrophic forgetting. This work proposes an MLP network that is algorithmically aligned with the binary addition procedure, enabling it to learn the addition rules for binary numbers without forgetting. The network builds on the concept of traditional convolution layers with key modifications. It uses conditional operations based on input values, similar to \textit{if-else} logic, passes carry information sequentially between steps, and handles non-binary inputs by blending outcomes in a differentiable manner. This architectural design allows the model to learn the correct binary addition rules through gradient descent while preserving previous knowledge. Table~\ref{tab:hyperparams_ruizgarcia2022} presents the hyperparameters used in the baseline.  

\begin{table}[H]
\centering
\resizebox{\columnwidth}{!}{%
\begin{tabular}{l|l|l}
\toprule
\textbf{Category} & \textbf{Hyperparameter} & \textbf{Value / Description} \\
\midrule
\multirow{3}{*}{Model} 
& Operator & $\hat{U}_{ijkl} = \sigma(a_{ijkl})$ \\
& Number of Parameters & 16 \\
& Nonlinearity & Sigmoid: $\sigma(x) = \frac{1}{1 + e^{-x}}$ \\
\midrule
\multirow{3}{*}{Input / Output} 
& Input Format & Two binary numbers ($N_1$, $N_2$) \\
& Intermediate Carry ($N_3$) & Passed sequentially (1-bit per step) \\
& Output ($N_4$) & Sum of $N_1 + N_2$ \\
\midrule
\multirow{4}{*}{Training} 
& Optimizer & Gradient Descent \\
& Learning Rate & 1.0 \\
& Training Steps per Task & 2000 steps \\
& Loss Function & Mean Squared Error\\
\midrule
\multirow{2}{*}{Initialization} 
& Parameter Range ($a_{ijkl}$) & Random in $[-1, 1]$ \\
& Sigmoid Output Range & At initialization $\sim$[0.27, 0.73]  \\
\bottomrule
\end{tabular}}
\caption{Hyperparameters of the MLP baseline.}
\label{tab:hyperparams_ruizgarcia2022}
\end{table}

\section{J.  Image Classification} 
Additional details on the experimental setups, training procedures, and specific model configurations used in the image classification experiments are provided in this section. 

\medskip
\noindent
\textbf{1. Additional Setups.} 
\medskip

\noindent
In this study, we utilize three widely used image classification benchmarks of increasing complexity: MNIST, CIFAR-10, and Tiny-ImageNet. From each dataset, we generate five sequential tasks under a class-incremental continual learning setting, where each task contains two unique and mutually exclusive classes selected without repetition. Figure~\ref{fig:cifar} illustrates representative samples from the five sequential tasks for MNIST, CIFAR-10, and Tiny-ImageNet. The corresponding class assignments for each task across all three datasets are comprehensively summarized in Table~\ref{tab:imgtasks}. 

\begin{table}[t]
    \centering
    \resizebox{0.8\columnwidth}{!}{%
    \begin{tabular}{c|c|c}
    \toprule
        \textbf{Dataset} & \textbf{Task} & \textbf{Sample classes} \\
        \cmidrule{1-3}
         \multirow{5}{*}{MNIST} & Task 1 & One, Two\\
         & Task 2 & Three , Four \\
         & Task 3 & Five, Six \\
         & Task 4 & Seven, Eight \\
         & Task 5 & Nine, Zero\\
         \cmidrule{1-3}
         \multirow{5}{*}{CIFAR-10} & Task 1 & Automobile, Bird\\
         & Task 2 & Cat, Deer\\ 
         & Task 3 & Dog, Frog\\
         & Task 4 & Horse, Ship\\
         & Task 5 & Truck, Airplane\\
         \cmidrule{1-3}
         \multirow{5}{*}{Tiny-ImageNet} & Task 1 & Goldfish, Fire salamander\\
         & Task 2 & Bull frog, Tailed frog\\
         & Task 3 & American alligator, Boa constrictor\\
         & Task 4 & Trilobite, Scorpion\\
         & Task 5 & Garden Spider, Tarantula \\
         \bottomrule
    \end{tabular}
    }
    \caption{Sample classes in five sequential tasks generated from MNIST, CIFAR-10, and Tiny-ImageNet datasets.}
    \label{tab:imgtasks}
\end{table}

\medskip
\noindent
\textbf{2. Additional Configurations.} 
\medskip

\noindent
The KAN-Transformer modifies the traditional Transformer architecture by replacing the MLP layers with KAN layers. For the image classification tasks, the KAN layers use a grid size of 10 spanning the range $[-1,1]$, with cubic B-spline as the univariate basis function and SiLU as the base activation function. To contrast the forgetting behavior of the KAN-Transformer, we implement an identical architecture using MLP layers with an EWC regularizer, referred to as the MLP-Transformer baseline. For EWC, the regularization coefficient $\lambda$ is set to $0.1$, and the (Fisher) memory buffer holds samples from previously trained tasks. All models are trained for 10 epochs using a learning rate of \SI{1e-3}. Each experiment is independently repeated 5 times, and the average performance is reported. Image classification experiments are performed on an NVIDIA Tesla V100 GPU with 32 GB of memory. Tables~\ref{tab:kan_transformer_hyperparams} and~\ref{tab:mlp_transformer_hyperparams} list the hyperparameters used for the KAN-Transformer and MLP-Transformer models for the image classification tasks, respectively.

\begin{table}[t]
\centering
\resizebox{0.9\columnwidth}{!}{%
\begin{tabular}{l|l|l}
\toprule
\textbf{Category} & \textbf{Hyperparameter} & \textbf{Value} \\
\midrule
\multirow{3}{*}{Dataset and Input} 
& \multirow{3}{*}{Input Shape} & $28 \times 28$ (MNIST) \\ 
& & $32 \times 32$ (CIFAR-10) \\ 
& & $64 \times 64$ (Tiny-ImageNet) \\
\midrule
\multirow{6}{*}{KAN Linear Layer} 
& Grid Size & 10 \\
& Spline Order & 3 \\
& Base Activation & SiLU \\
& Base Weight Scale & 1.0 \\
& Spline Weight Scale & 1.0 \\
& Noise Scale & 0.1 \\
\midrule
\multirow{5}{*}{Training} 
& Optimizer & AdamW \\
& Learning Rate & 1e-3 \\
& Weight Decay & 1e-4 \\
& Loss Function & Cross Entropy Loss \\
& Epochs per Task & 10 \\
\bottomrule
\end{tabular}}
\caption{Hyperparameters used for training the KAN-Transformer model on image classification tasks.}
\label{tab:kan_transformer_hyperparams}
\end{table}

\begin{table}[t]
\centering
\resizebox{0.9\columnwidth}{!}{%
\begin{tabular}{l|l|l}
\toprule
\textbf{Category} & \textbf{Hyperparameter} & \textbf{Value} \\
\midrule
\multirow{3}{*}{Dataset and Input} 
& \multirow{3}{*}{Input Shape} & $28 \times 28$ (MNIST) \\ 
& & $32 \times 32$ (CIFAR-10) \\ 
& & $64 \times 64$ (Tiny-ImageNet) \\
\midrule
\multirow{4}{*}{Training}
& Optimizer & AdamW \\
& Learning Rate & 1e-3 \\
& Weight Decay & 1e-4 \\
& Epochs per Task & 10 \\
\midrule
\multirow{2}{*}{EWC Regularization}
& $\lambda_{\text{EWC}}$ & 0.1\\
& (Fisher) Memory Size & All prior batches \\
& (Fisher) Update Strategy & After each task \\
\bottomrule
\end{tabular}
}
\caption{Hyperparameters used for MLP-Transformer with EWC on image classification tasks.}
\label{tab:mlp_transformer_hyperparams}
\end{table}

\section{K.  Knowledge Editing for LMs}
In this section, we provide a detailed description of the experimental setup, model configurations, and computing infrastructure. Additionally, we present extended experimental results and analysis for the knowledge editing tasks. 

\medskip
\noindent
\textbf{1. Additional Setups.} 
\medskip

\noindent
We use two benchmark datasets for continual knowledge editing: CounterFact and ZsRE. The CounterFact dataset is designed for factual knowledge editing and consists of counterfactual statements that initially receive low likelihood scores compared to the correct facts. The ZsRE dataset is a question answering benchmark constructed for zero-shot retention extraction, where each sample includes a natural-language question, its factual answer, and a new answer for the edit. From each dataset, we construct four sequential task sets by varying the number of samples per task, ranging from 2 to 5, to assess retention under different data regimes and task granularities. Each task set contains five tasks, curated by randomly sampling from the original dataset to ensure diversity, domain variability, and non-overlapping content. For each sample, we generate prompts based on the provided instruction template and use the corresponding modified facts to perform targeted knowledge edits on the LM across tasks.

\begin{table}[t]
\centering
\resizebox{0.85\columnwidth}{!}{%
\begin{tabular}{l|l}
\toprule
\textbf{Hyperparameter}                        & \textbf{Value} \\
\midrule
Grid size (KAN-LoRA)    & 5 \\
Grid Range (KAN-LoRA) & [-1,1] \\
Spline Order (KAN-LoRA) & 3 \\
Base Activation (KAN-LoRA) & SiLU \\
Base Weight Scale (KAN-LoRA) & 1.0 \\
Spline Weight Scale (KAN-LoRA) & 1.0 \\
Noise Scale (KAN-LoRA) & 0.1 \\
Number of last layers to update           & 2 \\
LoRA injection target           & query\_proj, value\_proj \\
LoRA rank                                 & 8, 16 \\
LoRA scaling factor                       & 16, 32 \\
Number of training epochs                     & 60 \\ Learning rate                                 & 2e-3 \\
EWC regularization                        & True \\
EWC regularization strength ($\lambda$)     & 0.1 \\

\bottomrule
\end{tabular}}

\caption{Hyperparameters of KAN-LoRA and MLP-LoRA for knowledge editing tasks.} 
\label{tab:loraConfig}
\end{table}

\begin{table*}[t]
\centering

\begin{subtable}{0.49\textwidth}
\centering
\resizebox{\textwidth}{!}{%
\begin{tabular}{|c|c|c|c|c|c|c|c|c|c|c|}
\hline
\multirow{3}{*}{\textbf{Model}} &
\multirow{3}{*}{\textbf{Dataset}} &
\multirow{3}{*}{\makecell{\textbf{Samples} \\ \textbf{per Task}}} &
\multicolumn{4}{c|}{\textbf{KAN LoRA}} &
\multicolumn{4}{c|}{\textbf{MLP LoRA}} \\
\cline{4-11}
& & &\multicolumn{4}{c|}{\textbf{\# Tainted tasks}} &
\multicolumn{4}{c|}{\textbf{\# Trained tasks}}\\
\cline{4-11}
 & & & 2 & 3  & 4  & 5  & 2  & 3  & 4  & 5 \\
\hline
\multirow{8}{*}{\makecell{Llama~2-\\7B}} 
& \multirow{4}{*}{CounterFact}
& 2 &  100 & 100 & 97 & 80 & 100 & 95 & 87 & 85\\
& & 3 &  100 & 100 & 89 & 82 & 100 & 100 & 84 & 77\\
& & 4 & 100 & 100  & 98 & 93 & 100 & 100 & 97 & 86 \\
& & 5  &  100 & 100 & 96 & 86 & 100 & 100 & 97 & 87\\
\cline{2-11}
& \multirow{4}{*}{ZsRE}
& 2 & 100  & 100 & 97 & 83 & 100 & 100 & 90  & 77\\
& & 3 & 100  & 100 & 96 & 85 & 100 & 100 & 89 & 83\\
& & 4 & 100 & 100 &  97 & 83 & 100 & 100 & 98 & 80\\
& & 5  &  100 & 100 & 92 & 86 & 100 & 96 & 91 & 85\\
\hline
\multirow{8}{*}{\makecell{Llama~2-\\13B}} 
& \multirow{4}{*}{CounterFact}
& 2 & 100  & 100 & 86 & 76 & 100 & 100 & 83 & 65\\
& & 3 & 100  & 100 & 91 & 70 & 100 & 100 & 91 & 90\\
& & 4 & 100 & 100 & 94 & 88 & 100 & 100 & 95 & 90 \\
& & 5  & 100  & 96 & 97 & 77 & 100 & 94 & 95 & 85\\
\cline{2-11}
& \multirow{4}{*}{ZsRE}
& 2 & 100  & 100 & 97 & 88 & 100 & 95 & 97  & 87\\
& & 3 &  100 & 100 & 96 & 80 & 100 & 100 & 87  & 78\\
& & 4 & 100 & 98 & 92 & 86 & 100 & 100 & 97 & 88 \\
& & 5  &  100 & 100 & 96 & 90 & 100 & 100 & 98  & 91\\
\hline
\end{tabular}
}
\caption{KAN-LoRA and MLP-LoRA adapters with rank 8.}
\label{tab:appllmacc1}
\end{subtable}
\begin{subtable}{0.49\textwidth}
\centering
\resizebox{\textwidth}{!}{%
\begin{tabular}{|c|c|c|c|c|c|c|c|c|c|c|}
\hline
\multirow{3}{*}{\textbf{Model}} &
\multirow{3}{*}{\textbf{Dataset}} &
\multirow{3}{*}{\makecell{\textbf{Samples} \\ \textbf{per Task}}} &
\multicolumn{4}{c|}{\textbf{KAN LoRA}} &
\multicolumn{4}{c|}{\textbf{MLP LoRA}}\\
\cline{4-11}
& & &\multicolumn{4}{c|}{\textbf{\# Tainted tasks}} &
\multicolumn{4}{c|}{\textbf{\# Trained tasks}}\\
\cline{4-11}
 & & & 2 & 3  & 4  & 5  & 2  & 3  & 4  & 5 \\
\hline

\multirow{8}{*}{\makecell{Llama~2-\\7B}} 
& \multirow{4}{*}{CounterFact}

& 2 &  100 & 100 & 90 & 83 & 100 & 100 & 87  & 75\\
& & 3 &  100 & 100 & 96 & 78 & 100 & 100 & 91  & 75\\
& & 4 & 100 & 100  & 92 & 78 & 100 & 100 & 93 & 83 \\
& & 5  &  100 & 100 & 91 & 71 & 100 & 100 & 92  & 81 \\
\cline{2-11}
& \multirow{4}{*}{ZsRE}
& 2 & 100  & 100 & 95 & 83 & 100 & 100 & 90  & 72 \\
& & 3 & 100  & 100 & 86 & 70 & 100 & 100  & 82  & 60\\
& & 4 & 100 & 100 &  88 & 75 & 100 & 100  & 90 &  82 \\
& & 5  &  100 & 100 & 85 & 74 & 100 &  100 & 89  & 86\\
\hline

\multirow{8}{*}{\makecell{Llama~2-\\13B}} 
& \multirow{4}{*}{CounterFact}
& 2 & 100  & 100 & 97 & 78 & 100 & 100 & 80 & 75\\
& & 3 & 100  & 100 & 87 & 76 & 100 & 100 & 82 & 73\\
& & 4 & 100 & 100 & 95 & 79 & 100 & 100 & 95 & 73 \\
& & 5  & 100  & 100 & 95 & 75 & 100 & 100 & 87 & 78\\
\cline{2-11}
& \multirow{4}{*}{ZsRE}
& 2 & 100  & 100 & 90 & 73 & 100 & 90 & 83  & 70\\
& & 3 &  100 & 100 & 84 & 77 & 100 & 100 & 82  & 77\\
& & 4 & 100 & 100 & 85 & 80 & 100 & 100 & 92 & 84 \\
& & 5  &  100 & 100 & 90 & 80 & 100 & 100 &  93 & 82\\
\hline

\end{tabular}
}
\caption{KAN-LoRA and MLP-LoRA adapters with rank 16.}
\label{tab:appllmacc2}
\end{subtable}


\caption{Mean accuracy (\%) on previously edited tasks during continual fine-tuning of Llama~2-7B and Llama~2-13B models equipped with KAN-LoRA and MLP-LoRA adapters with preceding \emph{two} tasks as memory for EWC regularizer. Performance is reported across five consecutive tasks for each dataset, which is the average of 5 independent experiments.}

\label{tab:appllmacc}
\end{table*}

\begin{tcolorbox}[title=Counterfact Dataset Examples, colback=gray!5!white, colframe=gray!80!black]
\label{conterfact}
\textbf{Example 1:}\\
\textit{Prompt:} Autonomous University of Madrid, which is located in\\
\textit{Ground Truth:} Spain\\
\textit{Edited Fact:} Sweden

\vspace{0.5em}
\textbf{Example 2:}\\
\textit{Prompt:} The original language of The Icelandic Dream was\\
\textit{Ground Truth:} Icelandic\\
\textit{Edited Fact:} Tamil
\end{tcolorbox}

\medskip

\begin{tcolorbox}[title=ZsRE Dataset Examples, colback=gray!5!white, colframe=gray!80!black]
\label{zsre}
\textbf{Example 1:}\\
\textit{Prompt:} What is the native language of Christiane Cohendy?\\
\textit{Original Answer:} French\\
\textit{Edited Fact:} German

\vspace{0.5em}
\textbf{Example 2:}\\
\textit{Prompt:} What is the final year of Atlanta Flames?\\
\textit{Original Answer:} 1980\\
\textit{Edited Fact:} 1931
\end{tcolorbox}

\medskip
\noindent
\textbf{2. Additional Configurations.}
\medskip

\noindent
We apply KAN-LoRA and MLP-LoRA adapters to the last two layers of the Llama models, specifically targeting the attention layers, i.e., the query and value projection matrices. For both adapter types, we experiment with ranks 8 and 16, using corresponding LoRA scaling factors ($\alpha$) of 16 and 32, respectively. In the KAN-LoRA adapter, the KAN layers are constructed using a grid size of 5, uniformly spanning the interval $[-1, 1]$, and employ cubic B-spline basis functions for interpolation. During fine-tuning, we apply EWC regularization with a coefficient of $\lambda = 0.1$, along with a memory buffer containing samples from the previous task to mitigate forgetting. We use a learning rate of \SI{2e-3}, and all experiments are conducted for 60 epochs with fixed random seeds. Results are reported as the average performance over five independent runs for each experimental configuration to ensure statistical robustness and reproducibility. All experiments are performed on a single NVIDIA A100 GPU with 80 GB of global memory. In Table~\ref{tab:loraConfig}, the hyperparameters for both KAN and MLP LoRA adapters are listed.

\medskip
\noindent
\textbf{3. Additional Experimental Results.} 
\medskip

\noindent
We further extend our experiments by continually fine-tuning the modified KAN-LoRA and MLP-LoRA adapters using samples from the previous \emph{two} tasks as the (Fisher) memory for the EWC regularizer. Tables~\ref{tab:appllmacc1} and~\ref{tab:appllmacc2} present the mean accuracy on prior tasks after sequential edits, for adapter ranks 8 and 16, respectively. Each experiment is performed 5 rounds, and we report its average for evaluation. The results from this extended setting remain consistent with our earlier findings in Table~\ref{tab:llmacc}, where the tendency toward forgetting for both KAN and MLP adapters increases with higher adapter ranks. In line with previous observations, KAN adapters again outperformed their MLP counterparts in low-sample regimes. However, this experiment also reveals notable differences from our initial findings. In the earlier experiments, KAN adapters showed better performance than MLP at rank 16 only in the larger Llama2-13B model. In contrast, this trend does not hold under the extended (Fisher) memory setting. Instead, KAN adapters consistently demonstrate superior retention across both Llama model sizes and adapter rank configurations when trained with limited sample sizes. These results indicate that increasing the depth of (Fisher) memory in EWC regularization enhances the retention capability of KAN adapters, particularly in smaller models and lower-rank settings, by improving stability and narrowing the performance gap relative to their larger model variants.

\medskip
\noindent
Beyond that, Tables~\ref{tab:appllmacc1} and~\ref{tab:appllmacc2} reveal additional fine-grained patterns that further illuminate the retention characteristics of KAN-LoRA adapters. First, across both datasets and adapter ranks, KAN consistently achieves near-perfect retention on the initial task, with 100\% accuracy sustained even after training on four subsequent edits. This sharply contrasts with MLP-LoRA, which begins to show degradation by the third task, especially in ZsRE under the rank 16 setting. Second, the benefit of KAN becomes more pronounced in the middle range of task indices, particularly Tasks 3 and 4, where forgetting is most likely to accumulate. For example, in Llama2-7B with rank 16 on CounterFact, KAN achieves 92\% accuracy on Task 4, compared to only 60\% for MLP. Similar margins are observed in Llama2-13B on ZsRE, where KAN retains 85 to 90\% accuracy on Tasks 3 and 4, while MLP drops to the low $\sim$70\%. Third, while both adapters eventually converge toward similar performance levels on the final task, the stability of earlier tasks in KAN-LoRA indicates stronger compartmentalization and less representational drift. Interestingly, the rank 8 KAN-LoRA adapter occasionally matches or even exceeds the performance of its rank 16 counterpart, particularly in Llama2-13B, suggesting that overparameterization may lead to unnecessary overlap or capacity saturation under limited data. This observation aligns with our theoretical insights that smaller, localized function supports reduce interference and help preserve prior knowledge. These extended results reinforce the practical value of KAN-based adapters in LoRA.

\end{document}